\documentclass[10pt,twocolumn,letterpaper]{article}
\usepackage[pagenumbers]{cvpr}
\usepackage[numbers,sort&compress]{natbib}

\usepackage[hyphens]{url}
\usepackage{graphicx}
\usepackage{amsmath,amssymb}
\usepackage{booktabs}
\usepackage{array}
\usepackage{multirow}
\usepackage{longtable}
\usepackage{float}
\usepackage{placeins}
\usepackage{tikz}
\usetikzlibrary{arrows.meta,calc,positioning,shapes.geometric}
\usepackage[pagebackref]{hyperref}
\hypersetup{hidelinks}

\title{World Model Control by Trajectory Reachability Models:\\ From Latent Distance to Task Progress}
\author{%
\small Liangyu Li\textsuperscript{1}, Shengzhi Wang\textsuperscript{1}, Libin Qiu\textsuperscript{2}, Mingliang Xiong\textsuperscript{1},
Wen Fang\textsuperscript{1}, Mingqing Liu\textsuperscript{1}, Jun Yang\textsuperscript{3}, Qingwen Liu\textsuperscript{1,*}\\[1pt]
\small \textsuperscript{1}Tongji University \quad \textsuperscript{2}Alibaba Group \quad \textsuperscript{3}Istituto Italiano di Tecnologia\\[1pt]
\footnotesize Liangyu Li: \href{mailto:2550703@tongji.edu.cn}{2550703@tongji.edu.cn}\quad
\textsuperscript{*}Corresponding author: Qingwen Liu (\href{mailto:qliu@tongji.edu.cn}{qliu@tongji.edu.cn})
}
\hypersetup{pdftitle={World Model Control by Trajectory Reachability Models: From Latent Distance to Task Progress},
pdfauthor={Liangyu Li, Shengzhi Wang, Libin Qiu, Mingliang Xiong, Wen Fang, Mingqing Liu, Jun Yang, Qingwen Liu}}

\newcommand{\z}{\mathbf{z}}
\newcommand{\zg}{\mathbf{z}_g}
\newcommand{\aseq}{\mathbf{a}_{0:H-1}}
\newcolumntype{L}[1]{>{\raggedright\arraybackslash}p{#1}}
\newcommand{\cem}{\mbox{\normalfont\scshape CEM}}
\newcommand{\tworoom}{\mbox{\normalfont\scshape TwoRoom}}
\newcommand{\pusht}{\mbox{\normalfont\scshape PushT}}
\newcommand{\lewm}{\mbox{\normalfont\scshape LeWM}}
\newcommand{\trm}{\mbox{\normalfont\scshape TRM}}
\newcommand{\pldm}{\mbox{\normalfont\scshape PLDM}}

\newcommand{\sasc}{\mbox{\normalfont\scshape SASC}}

\newcommand{\mlp}{\mbox{\normalfont\scshape MLP}}
\newcommand{\leworldmodel}{LeWorldModel}

\newcommand{\adamw}{AdamW}
\newcommand{\silu}{SiLU}
\newcommand{\softplus}{Softplus}
\newcommand{\smoothlone}{Smooth-L1}

\begin{document}
\maketitle

\begin{abstract}
A world model can encode the information needed for a task while its planner still selects the wrong actions. We study this discrepancy and find that, when a controller uses Euclidean latent distance as its terminal cost, directions carrying critical task information can contribute only a tiny fraction of the total distance. Accurately encoded information can therefore have little influence on candidate ranking. In TwoRoom, the position-related subspace contributes only 0.5--0.7\% of terminal latent squared distance, yet planning with this subspace alone raises success from 1.7\% to 90.8\%. Motivated by this finding, we introduce a trajectory reachability model (TRM) that learns a pairwise state cost from temporal relationships in logged trajectories and scores predicted endpoints against a goal, keeping the world model and planning algorithm fixed. On the main TwoRoom evaluation with all evaluation-source episodes excluded from pair training, TRM improves LeWorldModel from 7.0\% to 96.7\% success; on PLDM, it improves success from 32.7\% to 84.0\%. Shared-candidate ranking analysis, subspace interventions, and temporal-horizon ablations show that encoding control-relevant information and using it through a default distance are different problems. TRM provides a learned scoring interface between them.
\end{abstract}

\section{Introduction}
World models let agents learn environmental representations from visual observations, predict the consequences of candidate actions, and decide what to do next. A common approach encodes the current and goal observations into a latent space, predicts the latent state after an action sequence, and selects the plan closest to the goal. This turns visual control into distance optimization in latent space~\citep{chua2018pets,hafner2020dreamer,zhou2024dinowm,maes2026leworldmodel}.

This approach makes an easily overlooked assumption: states that are closer in representation space should also be closer to the goal for control. Learning to represent an environment does not automatically satisfy this assumption. A representation may preserve position, appearance, and other variation simultaneously. Summing errors across all directions can allow the remaining variation to overwhelm the few directions that determine an action~\citep{gelada2019deepmdp,zhang2021invariant}. The planner may then fail to advance along the directions needed to complete the task.

We observe this discrepancy in TwoRoom visual navigation. Two rooms are connected by a doorway: the agent cannot pass through the wall and must find the doorway route. LeWorldModel's latent representation permits almost exact linear recovery of position, with test $R^2\approx0.998$. Yet the controller using raw latent distance achieves only 1.7\% success on a distance-controlled evaluation. Replacing its terminal score with position distance decoded from the same predicted latent state raises success to 94.2\%.

We further localize the mechanism. The latent subspace used to recover position contributes only 0.5--0.7\% of total terminal distance. Planning with this subspace alone reaches 90.8\% success, whereas its orthogonal residual remains at 1.7\%. We view this as a common problem in this class of models: the default Euclidean distance assigns too little decision weight to task-relevant information. Scoring the same candidate actions in latent space and task space further reveals almost no agreement between their rankings.

This finding motivates our design. Rather than requiring a pretrained representation's Euclidean geometry to be naturally suited to control, we learn the score connecting representation to planning. Our trajectory reachability model (\trm{}) trains a lightweight cost head on temporal gaps between pairs of states in logged trajectories. Trajectories provide state connections that actually occurred in the environment; sampling across their full temporal span teaches the head longer-range relationships needed for planning, rather than only local similarity.

\begin{figure*}[!t]
\centering
\includegraphics[width=\linewidth]{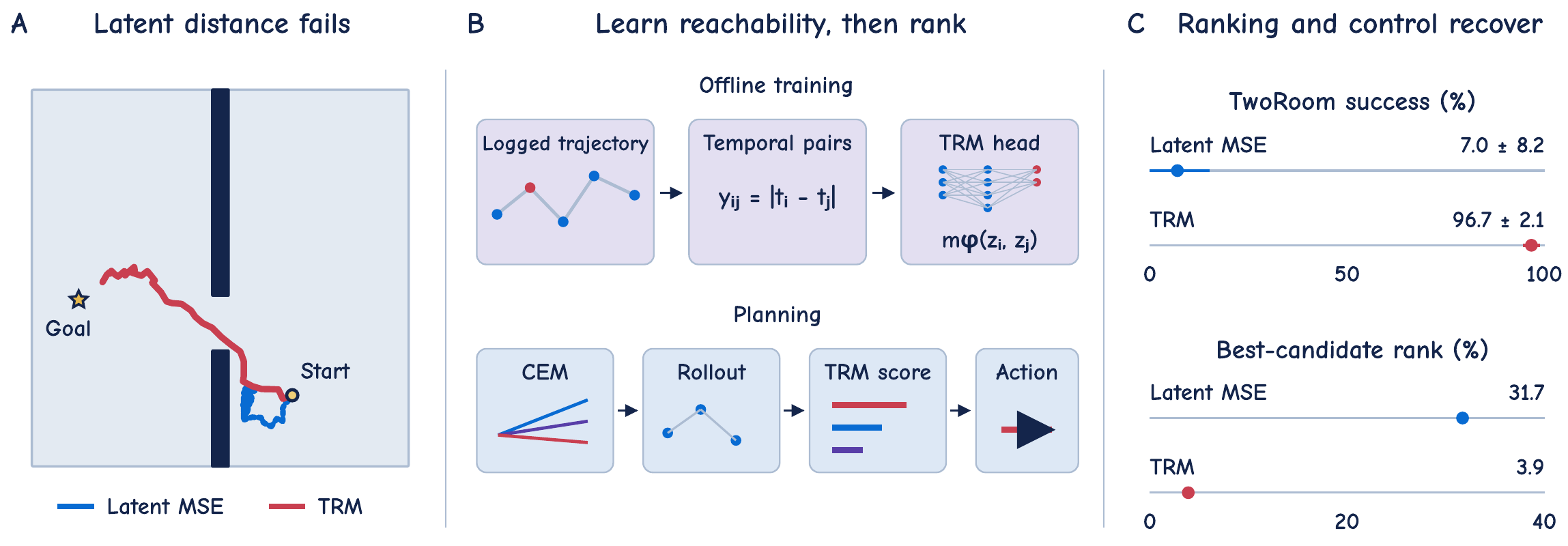}
\caption{Overview of the terminal ranking failure and the \trm{} repair. (A) Raw latent MSE drives \cem{} toward an unsuccessful plan, whereas \trm{} follows the doorway route. (B) Temporal pairs train the \trm{} head, which changes only terminal ranking under a fixed world model and controller. (C) The main high-distance control comparison improves success from $7.0{\pm}8.2$\% to $96.7{\pm}2.1$\%.}
\label{fig:trm_overview}
\end{figure*}

Our contributions form three connected parts:
\begin{enumerate}
\item \textbf{Identifying a scoring mismatch.} Position decoding, subspace interventions, and shared-candidate rankings show that task information can be present in a representation yet underweighted by its default terminal distance.
\item \textbf{Learning a trajectory-driven scoring interface.} With the world model frozen, TRM learns from full-horizon trajectory relationships to use the representation for candidate action selection.
\item \textbf{Connecting mechanism to control.} Matched-budget evaluations, shuffled labels, temporal-horizon ablations, coverage tests, and model and task comparisons establish the sources and scope of the gains.
\end{enumerate}

\section{Related Work}

\paragraph{From predictive representations to action selection.}
World models support control by predicting the consequences of actions, from probabilistic dynamics~\citep{deisenroth2011pilco,chua2018pets,janner2019mbpo} and visual foresight~\citep{ebert2018visualforesight} to latent dynamics~\citep{watter2015embed,hafner2019planet}. Learned models can train policies through imagined experience~\citep{ha2018worldmodels,hafner2020dreamer,hansen2022tdmpc,hansen2024tdmpc2}, or score candidate action sequences against a visual goal, as in DINO-WM and LeWorldModel~\citep{zhou2024dinowm,maes2026leworldmodel}. The latter setting exposes a specific interface: a predictive representation must also provide a useful ordering of candidate endpoints. This connects to the objective mismatch between model learning and control~\citep{lambert2020objective}. Our investigation begins at this interface, using interventions on an already trained model to separate available state information from its influence on action selection.

\paragraph{Learning how far a goal is.}
Goal-conditioned value functions and reachability learning provide alternatives to visual similarity~\citep{schaul2015universal,andrychowicz2017hindsight,nair2018rig,eysenbach2021clearning,eysenbach2022contrastive,ma2023vip}. Graph-based and hierarchical methods organize reachable states for long-horizon decisions~\citep{nachum2018hiro,savinov2018semi,eysenbach2019sorb,zhang2021worldgraph}; temporal and quasimetric representations similarly encode behavioral relations~\citep{dayan1993successor,bae2024tldr,park2024metra,park2024hilp,myers2024learningtemporal,myers2025quasimetric}. MBOLD learns a functional distance through offline Q-learning and uses it to score model-predicted outcomes in visual MPC~\citep{tian2021mbold}. Its temporal-regression comparison makes time-pair sampling and label clipping particularly relevant controls for a temporal head. LEXA uses a world model to train goal-achieving policies in imagination~\citep{mendonca2021lexa}, while RePlan separates representation learning from temporal reachability estimation~\citep{qian2023replan}. These works establish learned behavioral distance as a planning tool. Our focus is the diagnosis and repair of terminal ranking in a frozen world model, with full-horizon temporal supervision and shared-candidate interventions that test how the learned score changes decisions.

\paragraph{Shaping world models for planning.}
Value-guided JEPA trains representations so that a distance or quasidistance approximates a goal-conditioned reaching value~\citep{destrade2025valueguided}. RC-aux adds multi-horizon prediction and budget-conditioned reachability supervision to world-model training or continuation training~\citep{li2026predictive}. Both address the gap between prediction and plannability through learned behavioral structure. TRM studies a complementary intervention: the encoder and predictor remain frozen, and a post-hoc pairwise head supplies a scalar endpoint score. This distinction isolates what can be recovered from the existing representation without attributing gains to a newly trained world model. TEMPO also keeps the world model frozen, learning a temporal embedding whose distance is blended with the original latent cost~\citep{moukheiber2026tempo}. Its released workflow refines that embedding using the planner's own trajectories. TRM directly regresses a pairwise scalar from logged temporal pairs; the shared-candidate and rowspace experiments additionally test which information and rankings account for the control change.

\paragraph{The score and the search process.}
CEM~\citep{rubinstein1999crossentropy,pinneri2021icem} and path-integral control~\citep{williams2017mppi} repeatedly rank candidate action sequences. Better optimization, train--test alignment~\citep{parthasarathy2025closing}, and learned imagination priors~\citep{wang2026prism} can improve adjacent parts of this process. We hold the world model, proposal mechanism, and search budget fixed when comparing terminal costs. Shared-candidate audits additionally separate ranking quality from the subsequent proposals and trajectories induced by a changed score.

\section{Why Does Encoded Information Not Become Correct Actions?}
\begin{figure}[!htbp]

\centering
\includegraphics[width=\linewidth]{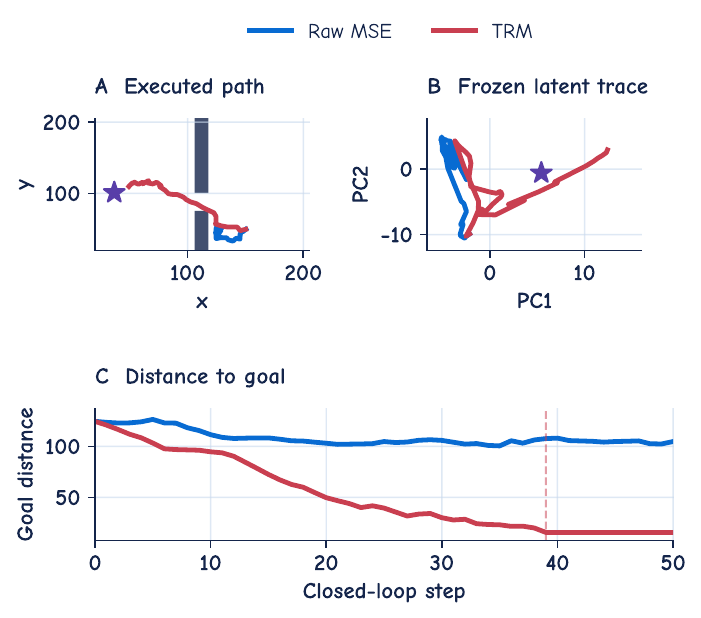}
\caption{Representative high-distance TwoRoom execution. Raw latent MSE stops near the wall, while temporal TRM follows the doorway route.}
\label{fig:latent_reachability_cartoon}

\end{figure}
\label{sec:diagnosis}
\subsection{Starting from a Navigation Failure}
We begin with a planning failure in TwoRoom visual navigation. The environment contains two rooms separated by a wall and connected by a doorway. Reaching a goal across the wall requires moving toward the doorway before entering the other room. On a distance-controlled evaluation with an action budget of 50, the original LeWorldModel controller succeeds on only 1.7\% of cases. We ask why the world model fails to support this apparently simple route decision.

The controller encodes an observation $o_t$ as $\z_t=f_\theta(o_t)$ and predicts a terminal latent state under an action sequence $\aseq$:
\begin{equation}
\hat{\z}_{t+H}=F_\theta^{(H)}(\z_t,\aseq).
\end{equation}
For goal latent $\zg=f_\theta(o_g)$, its terminal objective is
\begin{equation}
 c_{\mathrm{lat}}(\aseq)=\|\hat{\z}_{t+H}-\zg\|_2^2.
 \label{eq:latent-cost}
\end{equation}
The cross-entropy method (\cem{})~\citep{rubinstein1999crossentropy} samples action sequences, ranks them by this score, updates its search distribution, executes a receding-horizon prefix, and replans. The score therefore determines which predicted outcomes guide action selection.

\subsection{The Information Is Present but Barely Affects the Score}
The first question is whether the representation has lost the position information needed for navigation. We test this with a linear map from the latent state to XY position. Trained on 20,000 cached states, this map achieves test RMSE of approximately 1.8 pixels and $R^2=0.998$ across all three LeWorldModel seeds. Position is thus recoverable from the learned representation.

We next use this decoded position to score the same model's predicted terminal states. Keeping the world model and controller fixed, the decoded-XY terminal cost raises success from 1.7\% to 94.2\%. The evaluation does not supply oracle terminal coordinates: the score uses a learned linear map of the model's predicted latents.

To isolate how much this information contributes to the original distance, let the position map be $W\z+b$, with rowspace projector
\begin{equation}
P_W=W^\top(WW^\top)^\dagger W.
\end{equation}
Writing $\delta\z=\hat{\z}_{t+H}-\zg$, the terminal squared distance decomposes as
\begin{equation}
\|\delta\z\|_2^2=\|P_W\delta\z\|_2^2+\|(I-P_W)\delta\z\|_2^2.
\end{equation}
The position-related rowspace contributes only 0.5--0.7\% of total terminal latent distance. Nevertheless, using its component alone as the planning cost reaches 90.8\% success; using only the orthogonal residual remains at 1.7\%. The original score gives negligible influence to directions that can support successful control.

Long-horizon tasks may further exacerbate this issue. A single planning step typically advances only a small fraction of a long task, so candidate actions may differ little in task progress. When the resulting differences in latent-space scores are comparable in scale to those induced by task-irrelevant representation changes, raw latent distance may primarily reflect the latter and struggle to distinguish actions that better advance the task. We observe a pattern consistent with this interpretation in offline TwoRoom trajectories. Across the first ten steps of 8,302 trajectories, the squared norm of the one-step latent change projected onto the XY-related subspace accounts for only 0.0110--0.0136\% of the total change. Even after normalizing by dimensionality, the per-dimension squared change in this subspace is only 1.05--1.29\% of that in its orthogonal complement. Meanwhile, the ratio of one-step physical XY displacement to the current straight-line goal distance averages 6.51\% early in the trajectory, compared with 17.99\% late in the trajectory. These results show that early one-step progress is small relative to the remaining goal scale and that XY-related changes occupy a small share of one-step latent variation, supporting the above interpretation.

\begin{figure}[!htbp]
\centering
\includegraphics[width=\linewidth]{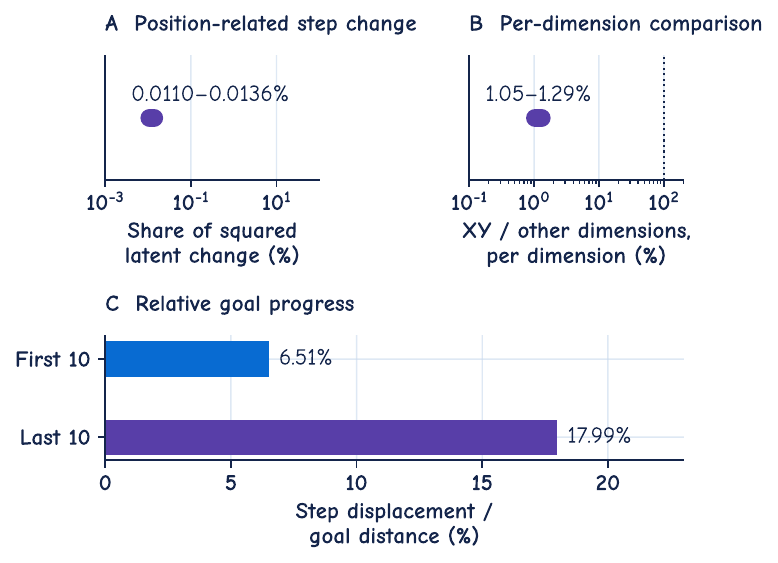}
\caption{One-step changes in offline TwoRoom trajectories. A: XY-rowspace share of squared latent change. B: per-dimension XY change relative to its orthogonal complement. C: step displacement relative to goal distance, early versus late; not directed progress.}
\label{fig:one_step_changes}
\end{figure}

\subsection{The Mismatch Appears in Candidate Rankings}
\begin{figure}[!htbp]
\centering
\includegraphics[width=\linewidth]{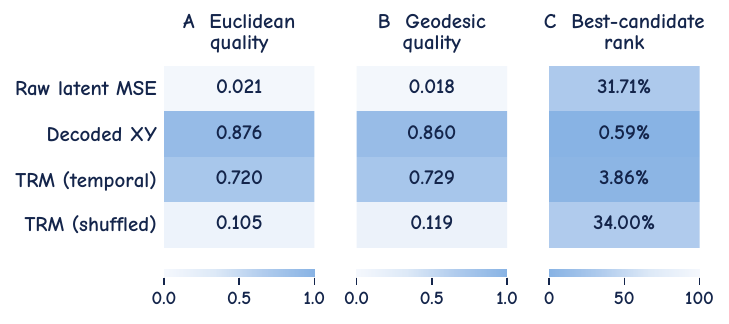}
\caption{Shared-candidate ranking on the high-distance TwoRoom evaluation (one seed). Each column uses its own scale. Higher correlations and lower best-candidate rank percentiles are better. Decoded XY uses diagnostic geometric information. Cells are aggregate metrics, not individual candidate outcomes.}
\label{fig:candidate_heatmap}
\end{figure}

We then compare costs on the same candidate action sequences using a selection audit with shared candidates (\sasc{}). A task-space diagnostic cost supplies the reference ordering; holding the candidates fixed separates ranking from differences in what the controller proposes. On the main TwoRoom audit for seed 3072, raw latent distance has geodesic rank correlation $\rho=0.018$ and assigns the oracle-best candidate a rank percentile of 31.71. Decoded XY improves these to 0.860 and 0.59, respectively (Figure~\ref{fig:candidate_heatmap}).

Increasing the raw-distance search budget to 1,000 samples, 20 CEM iterations, and top-$k$ 100 still gives only 2.5\% success. If search continues to rely on a mismatched score, it cannot identify physically feasible actions. Together, the interventions motivate learning a terminal score that gives task-relevant relationships enough influence over action selection.

\section{Method: Trajectory Reachability Model (\trm{})}

\subsection{Learning Actual State Connections Instead of Default Geometry}
We use the unlabeled temporal evolution in recorded data to learn a pairwise cost for planning while keeping the world model's encoder and predictor fixed. For states $(\z_i,\z_j)$ from the same trajectory, the label is
\begin{equation}
y_{ij}=|t_i-t_j|.
\end{equation}
This measures their temporal separation under the recorded behavior. Trajectories connect states through transitions that actually occurred, supplying supervision different from visual similarity: the head learns which states are nearby along the recorded behavior and which are far apart. TRM uses a nonnegative pairwise head $m_\phi(\z_i,\z_j)$ with features
\begin{equation}
[\z_i,\z_j,\z_i-\z_j,|\z_i-\z_j|].
\end{equation}

\subsection{Long-Range Planning Requires Long-Range Training Pairs}
If most training pairs are neighboring states, a head can distinguish local changes without learning the longer-range relationships between planning endpoints. The sampling rule is therefore part of the method. We sample an episode, choose a temporal separation over the full episode horizon, choose a valid start time, and train on pairs in random order. We train on 100,000 balanced pairs using Smooth-L1 loss on temporal labels scaled by 224. Balanced sampling over the full horizon prevents the head from being dominated by local neighbors. It also exposes the head to the temporal scale at which the terminal cost will rank \cem{} candidates. Table~\ref{tab:trm_ablation_main} gives the key comparison. In the seed 3072 diagnostic, broad random coverage over full episodes reaches 90.0\%, balanced pairs over full episodes reach 97.5\%, and a short-horizon head with maximum $\Delta=50$ reaches only 35.0\% even with 100,000 training pairs. The corresponding three-seed horizon comparison is reported in Table~\ref{tab:trm_ablation_main}.

\subsection{Using the New Score in the Original Planner}
\begin{figure}[!htbp]
\centering
\includegraphics[width=\linewidth]{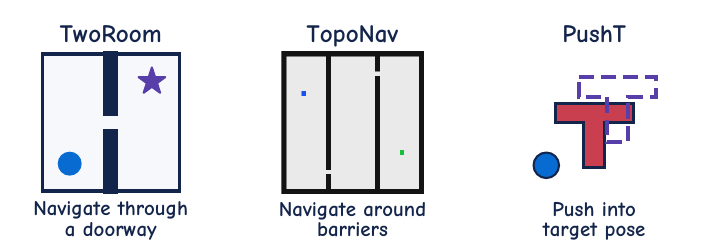}
\caption{Core diagnostic tasks. The scenes are illustrative task schematics, not evaluation screenshots; TopoNav markers illustrate a start and goal.}
\label{fig:evaluation_suite}
\end{figure}

At control time, \cem{} proposes candidates, the fixed world model rolls each candidate to a predicted terminal latent state $\hat{\z}^{(i)}_{t+H}$, and \trm{} computes the temporal terminal cost against the goal latent state.
\begin{equation}
    c_{\trm}(\aseq^{(i)})=m_\phi\!\left(\hat{\z}^{(i)}_{t+H},\zg\right).
\end{equation}
In settings dominated by reachability, we use $c_{\trm}$ as a replacement terminal cost. In settings where raw latent distance may preserve useful local physical information, we use the standardized hybrid cost below.
\begin{equation}
    c_{\mathrm{hyb}}=\operatorname{zscore}(c_{\mathrm{lat}})+\lambda\,\operatorname{zscore}(m_\phi).
    \label{eq:hybrid}
\end{equation}
Here $\operatorname{zscore}$ standardizes costs within the current candidate set, $\operatorname{zscore}(c_i)=(c_i-\mu_c)/(\sigma_c+\epsilon)$. The \pusht{} sweeps $\lambda$ in the supplement, whereas \tworoom{} primarily uses replacement costs.
\trm{} is a scalar endpoint terminal cost used by \cem{} for action selection. Figure~\ref{fig:evaluation_suite} summarizes the core diagnostic tasks and their evidence; the additional Reacher and Cube evaluations are described in Section~\ref{sec:experiments}.

\section{Experiments: Temporal Scoring for World-Model Control}
\label{sec:experiments}
We organize the experiments around three questions: how temporal scoring compares with existing control approaches, which properties of the score and its supervision explain the results, and under what task conditions those benefits persist. We first present navigation performance and a goal-conditioned policy comparison, then analyze candidate ranking and temporal supervision, and finally examine coverage, geometry, and continuous-control tasks.

\paragraph{Models.}
We study terminal scoring in \lewm{} and test its transfer to a fixed \pldm{} world model~\citep{maes2026stableworldmodel}.

\paragraph{Evaluation and reporting.}
We compare terminal scores with the world model, evaluation cases, and planning budget held fixed. Temporal heads learn from logged trajectories. Tables state the sample sizes and seed counts; complete protocols appear in Appendix~\ref{app:experimental_implementation}.

\begin{figure}[!htbp]
\centering
\includegraphics[width=\linewidth]{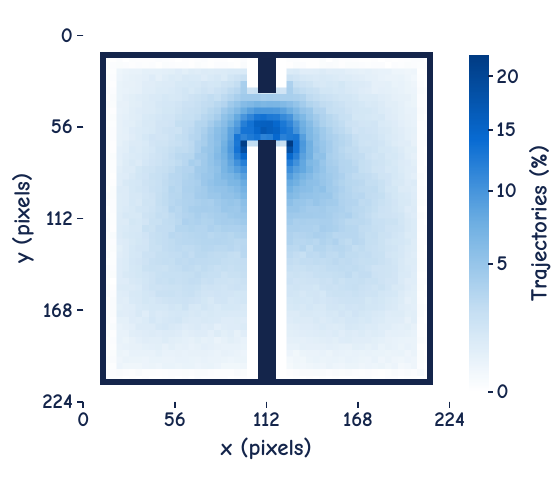}
\caption{TwoRoom training-source coverage. Percentage of source trajectories visiting each spatial cell. White indicates no visits; colors use a power scale.}
\label{fig:training_coverage}
\end{figure}

For TwoRoom, we draw temporal training pairs from a pool of 9,807 trajectories spanning both rooms and the connecting doorway (Figure~\ref{fig:training_coverage}).

\subsection{Control Performance and Method Comparisons}
\begin{figure}[!htbp]
\centering
\includegraphics[width=\linewidth]{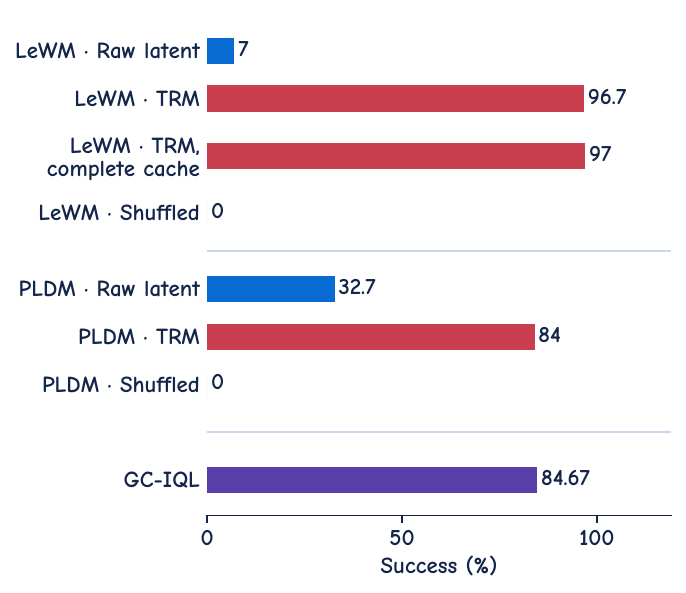}
\caption{TwoRoom control performance. GC-IQL learns from logged actions and differs from TRM in supervision and optimization budget. Complete per-seed uncertainty and failure statistics are retained in Tables~\ref{tab:hardn100} and~\ref{tab:gc_iql}.}
\label{fig:control_comparison}
\end{figure}

\paragraph{Long-range navigation across world models.}
We use the high-distance \tworoom{} evaluation to measure how the terminal score affects closed-loop navigation (Figure~\ref{fig:control_comparison}). Temporal \trm{} reaches 96.7\% mean success, compared with 7.0\% for raw \lewm{} latent MSE. The corresponding wrong-room and wall-stuck rates decrease to 2.0\% and 0.3\%. Shuffled temporal labels yield 0.0\% success, indicating that the learned temporal relation, rather than the additional head alone, provides the useful planning signal.

The same temporal scoring recipe also improves \pldm{} success from 32.7\% to 84.0\%.

\paragraph{Goal-conditioned policy comparison.}
On TwoRoom, temporal scoring reaches 96.67\% success, compared with 84.67\% for a goal-conditioned IQL policy~\citep{kostrikov2021iql} and 7.00\% for raw latent planning (Figure~\ref{fig:control_comparison}). These methods use the same evaluation cases and action budget. GC-IQL additionally learns from logged actions, whereas TRM scores endpoints within a frozen world-model planner; their supervision and optimization budgets differ.

\subsection{Mechanism and Temporal Supervision}
\paragraph{Candidate-ranking analysis.}
We use the shared-candidate analysis (\sasc{}) to measure how a terminal score ranks possible actions. Each selector scores the same action sequences with the same checkpoint and planning budget. We compare controller cost $c(i)$ with diagnostic task cost $c^\star(i)$, computed from known task-geometry rollouts in TwoRoom, exact grid dynamics in TopoNav, and simulator-executed outcomes in PushT. We report within-pool Spearman correlation, the oracle-best candidate's rank, and selected task distance. Higher correlation and lower rank or distance are better (Appendix~\ref{app:sasc_audit}).

\begin{figure}[!htbp]

\centering
\includegraphics[width=0.48\textwidth]{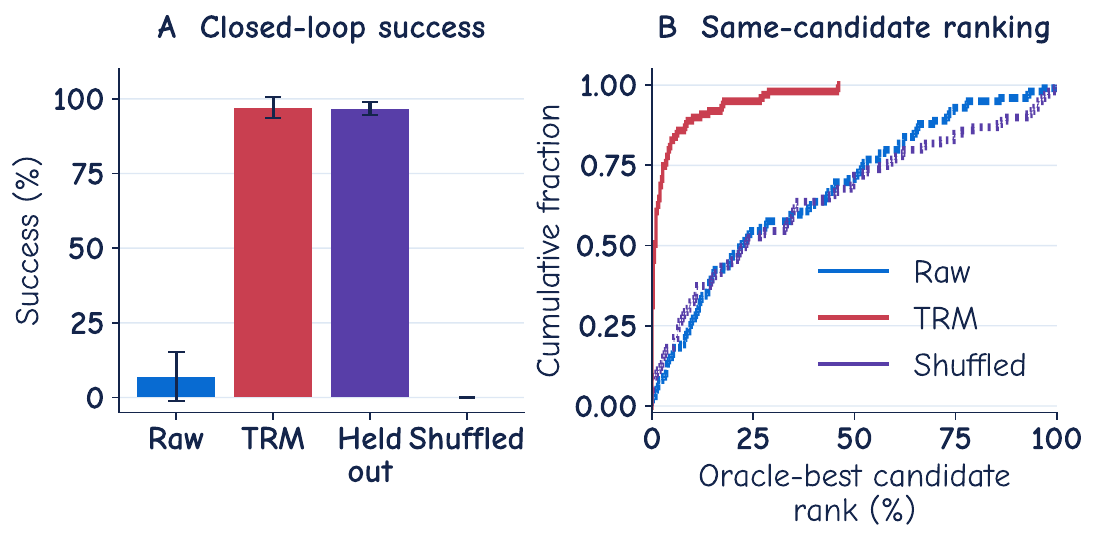}
\caption{Control and shared candidate ranking. (A) Success on the high distance \tworoom{} evaluation across three \lewm{} seeds. (B) Single-seed \sasc{} rank distributions on fixed candidates from the same evaluation, where left is better.}
\label{fig:control_ranking}
\end{figure}

TRM improves how the planner orders candidate actions: geodesic Spearman rises from 0.018 to 0.729, while the oracle-best candidate moves from rank percentile 31.71 to 3.86 (Figure~\ref{fig:candidate_heatmap}). Figure~\ref{fig:control_ranking} relates this ranking change to closed-loop success.

\paragraph{Candidate availability and ranking.}
We separate candidate availability from action selection. With 300 candidates, the available progress in TRM-guided pools is 127.83 pixels, compared with 38.17 pixels in raw-distance-guided pools. In a separate fixed-pool comparison, selection regret measures the actual distance gap between the selected action and the best action in the pool. On raw-distance-guided pools, this gap is 22.67 pixels for raw scoring and 13.02 pixels for TRM; on TRM-guided pools, it is 13.32 and 8.69 pixels, respectively. Appendix~\ref{app:candidate_decomposition} reports the two experiments separately.

\begin{figure}[!htbp]

\centering
\includegraphics[width=0.48\textwidth]{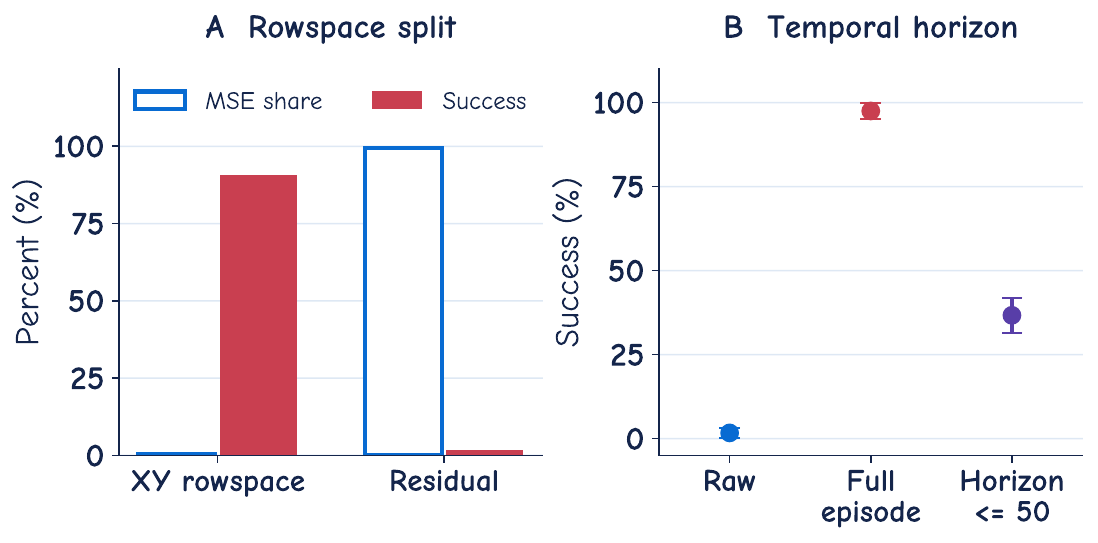}
\caption{Mechanism and horizon evidence. (A) Latent MSE share and control success for XY rowspace and residual costs. (B) Success in the distance controlled evaluation with an action budget of 50 under raw, full episode, and short horizon costs.}
\label{fig:mechanism_horizon}
\end{figure}

The subspace analysis in Section~\ref{sec:diagnosis} provides a complementary explanation: the position-relevant subspace contributes only 0.5--0.7\% of raw latent MSE, yet using it alone raises success from 1.7\% to 90.8\%. Figure~\ref{fig:mechanism_horizon}A summarizes this result. Together, the subspace and candidate-ranking analyses connect information retained by the model to the decisions induced by its score.

We further analyze the sensitivity of TRM's score to position-related directions. The XY subspace accounts for 3.49--4.64\% of the squared norm of its score gradient, approximately 3.4--4.5 times the isotropic baseline, indicating greater relative sensitivity of the learned score to these directions.

\begin{figure}[!htbp]
\centering
\includegraphics[width=\linewidth]{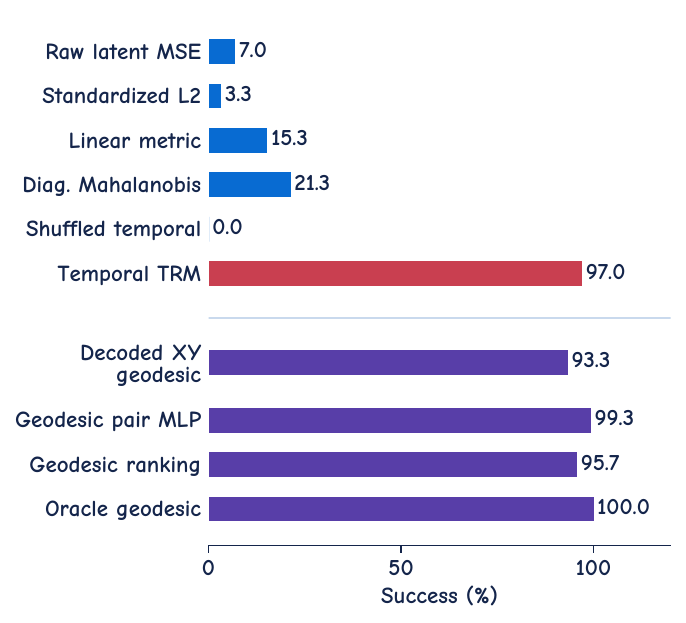}
\caption{Terminal-score comparisons on matched high-distance TwoRoom evaluations, averaged over three seeds. Full ranking and distance statistics appear in Table~\ref{tab:decisive_stress}.}
\label{fig:metric_comparison}
\end{figure}

\paragraph{Beyond latent rescaling.}
We compare latent rescaling, learned metrics, and geometry-based scores to identify the information needed for effective candidate selection (Figure~\ref{fig:metric_comparison}). Simple rescaling and linear metric adjustments remain insufficient for reliable navigation. Temporal supervision yields 97.0\% success without explicit geodesic labels, approaching methods that use task geometry or geometric supervision.

In recorded TwoRoom trajectories, temporal separation is strongly correlated with shortest feasible path distance: the overall Spearman correlation is 0.883, and it is 0.779 for state pairs whose paths pass through the doorway.

\paragraph{Long-range supervision.}
\label{sec:trm_ablation}

Long-range supervision improves long-range control. On the distance-controlled evaluation, full-episode temporal supervision reaches 97.5\% success, compared with 36.7\% for short-horizon supervision and 1.7\% for raw latent MSE (Table~\ref{tab:trm_ablation_main}; Figure~\ref{fig:mechanism_horizon}B). The temporal heads use the same training-pair count and planning budget, isolating the effect of temporal coverage.

\begin{table*}[!t]
\caption{Horizon comparison on the distance controlled evaluation with an action budget of 50 across three \lewm{} seeds. Both temporal heads use the same training-pair count. Broader coverage and pair count sweeps are single seed diagnostics in the supplementary material.}
\label{tab:trm_ablation_main}
\centering
\begin{tabular}{L{0.15\linewidth}L{0.24\linewidth}rrrr}
\toprule
Sampling & Temporal horizon & Run 1 & Run 2 & Run 3 & Mean$\pm$std \\
\midrule
Raw latent MSE & None & 2.5 & 2.5 & 0.0 & $1.7{\pm}1.4$ \\
Balanced temporal & Full episode & 97.5 & 100.0 & 95.0 & $97.5{\pm}2.5$ \\
Balanced temporal & $\Delta \le 50$ & 35.0 & 42.5 & 32.5 & $36.7{\pm}5.2$ \\
\bottomrule
\end{tabular}
\end{table*}

The temporal relationships themselves matter: shuffled labels yield 0.0\% success for both \lewm{} and \pldm{}. The single-seed ranking diagnostic also deteriorates (Figure~\ref{fig:candidate_heatmap}). These controls keep the architecture, data volume, and evaluation unchanged.

Additional supervision and implementation details are provided in Appendix~\ref{app:head_settings}.

\subsection{Task Conditions and Scope of Transfer}
\paragraph{Trajectory coverage and geometry.}
\label{sec:coverage_layout_stress}

TopoNav is a multi-room navigation task in which the agent must navigate around partition walls and pass through staggered openings to reach the goal. We use this task to analyze whether planning methods can find routes through obstacles, and how they perform when training trajectories cover key passages.

\begin{figure}[!htbp]
\centering
\includegraphics[width=\linewidth]{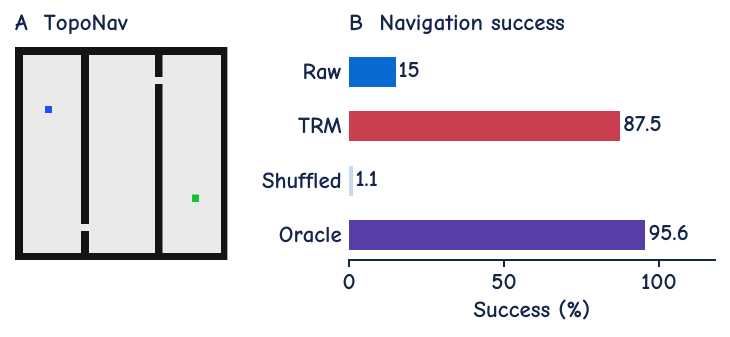}
\caption{TopoNav layout and navigation success. The map shows the task geometry, with illustrative start and goal markers. Protocol and results are in Table~\ref{tab:appendix_toponav}.}
\label{fig:toponav_results}
\end{figure}

\looseness=-1
TRM achieves 87.5\% success, compared with 15.0\% for raw latent distance (Figure~\ref{fig:toponav_results}). In the separate TwoRoom doorway-coverage diagnostic, success is 100.0\% with balanced doorway coverage and 0.0\% with limited doorway coverage.

\begin{table*}[!t]
\caption{Supporting evaluations under fixed controller budgets. Full settings are in the supplement.}
\label{tab:secondary_evidence}
\centering
\begin{tabular}{lll}
\toprule
Check & Main signal & Appendix evidence \\
\midrule
TopoNav & 87.5\% success & Table~\ref{tab:appendix_toponav} \\
TwoRoom doorway coverage & Rank 34.62 to 0.49 & Table~\ref{tab:appendix_coverage_layout} \\
\pusht{} boundary & Distance improves more than success & Appendix~\ref{app:pusht_details} \\
\bottomrule
\end{tabular}
\end{table*}

\paragraph{Continuous-control score transfer.}
Reacher and Cube test whether temporal scoring remains useful beyond navigation.

\begin{table}[t]
\caption{Temporal scoring beyond navigation. Success (\%) in Reacher and Cube. Width is the hidden width of the temporal head.}
\label{tab:new_continuous}
\centering
\begin{tabular}{lrr}
\toprule
Terminal cost & Reacher & Cube \\
\midrule
Raw latent & 88.67 & 6.67 \\
Temporal TRM, width 256 & 93.67 & 7.33 \\
Temporal TRM, width 1024 & 94.00 & 11.00 \\
\bottomrule
\end{tabular}
\end{table}

TRM improves performance in both Reacher and Cube: success increases from 88.67\% to a maximum of 94.00\% in Reacher, and from 6.67\% to a maximum of 11.00\% in Cube.

\paragraph{Online-policy references.}
\begin{table}[t]
\caption{Online-policy references. }
\label{tab:rl_reference}
\centering
\begin{tabular}{llr}
\toprule
Task & Policy & New success (\%) \\
\midrule
Reacher & Online PPO & 4.00 \\
Cube & Online PPO & 0.00 \\
\bottomrule
\end{tabular}
\end{table}

We further compare continuous-control performance with PPO, an online reinforcement learning method~\citep{schulman2017ppo}. PPO achieves new-success rates of 4.00\% and 0.00\% in Reacher and Cube, respectively (Table~\ref{tab:rl_reference}).

\paragraph{PushT.}
\pusht{} tests whether better candidate ranking translates to better contact control. We use the official \lewm{} checkpoint and expert demonstrations, varying goal offset to change task difficulty. All methods share the evaluation cases and action budget within each offset; the full settings are in Appendix~\ref{app:pusht_details}.

\begin{figure}[!htbp]
\centering
\includegraphics[width=\linewidth]{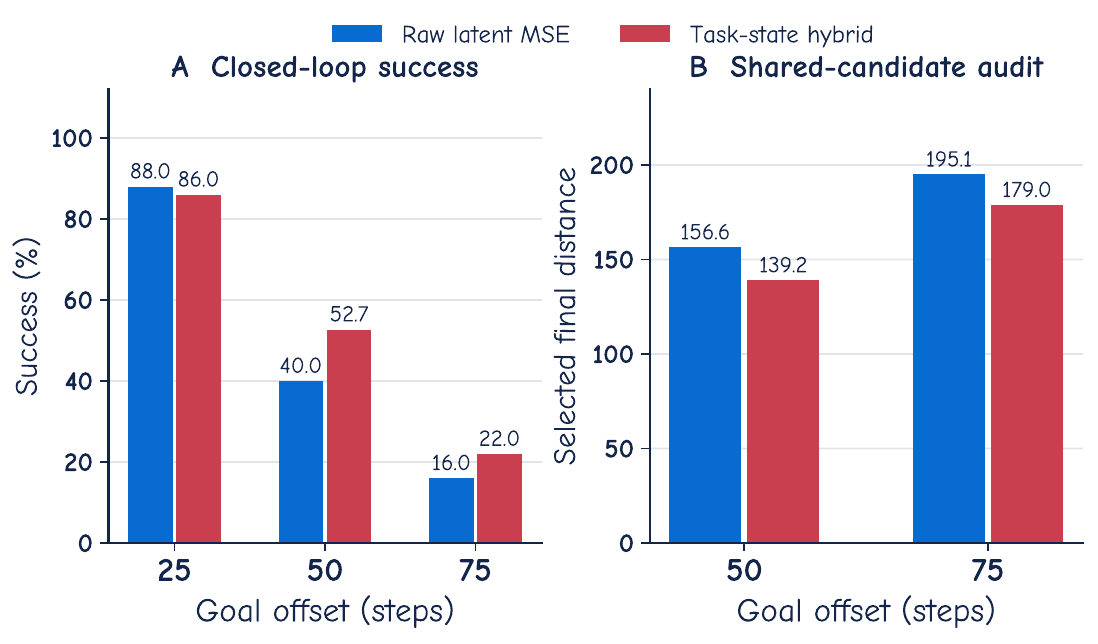}
\caption{\pusht{} boundary evidence. Closed loop success improves modestly, while \sasc{} shows larger reductions in selected final distance.}
\label{fig:pusht_boundary_main}
\end{figure}

\pusht{} tests \trm{} style task state heads as auxiliary ranking signals. At goal offset 25, raw latent planning is already near saturated at 88.0\%, and learned replacement or hybrid costs do not significantly improve it. At the harder goal offsets 50 and 75, task state heads strongly improve \sasc{} ranks and selected final distance, but closed loop success improves only modestly. The offset 50 comparison is raw 40.0\% versus task state hybrid 52.7\%, and the offset 75 comparison is raw 16.0\% versus task state hybrid 22.0\%. Figure~\ref{fig:pusht_boundary_main} summarizes this boundary. The learned task state cost is informative, while continuous contact manipulation still depends on rollout fidelity, contact execution, action search, and recovery. Detailed mechanism tables and \pusht{} sweeps are in the supplementary material.

\section{Scope and Limitations}

The strongest control gains are established in visual navigation. In \tworoom{} and TopoNav, temporal replacement works when candidate pools contain viable routes and logged trajectories cover connectivity-critical regions; rare doorway coverage fails, while balanced coverage restores ranking and success. \pusht{} is more continuous: raw latent cost retains useful physical signal, and learned task state heads work best as hybrid auxiliary costs.

\pusht{} improves candidate ranking more clearly than closed loop success. 

\section{Conclusion}
We identify a failure mode of world models: a model may well have learned the information needed to complete a task, yet the raw latent distance used for planning fails to select the correct actions. Through candidate-ranking and subspace analyses, we find that task-relevant directions may contribute too little to latent distance, making it difficult for the planner to identify actions that better advance the task. Based on this finding, we propose the Trajectory Reachability Model (TRM), which learns a planning-oriented terminal score from temporal relationships in trajectories. On navigation tasks including TwoRoom and TopoNav, TRM improves candidate ranking and closed-loop control, demonstrating one way to translate the representational capabilities of world models into planning capabilities.

\section*{Acknowledgments}
The authors used OpenAI ChatGPT and Codex for literature search, experimental code development, and drafting or editing text, figures, and LaTeX. The authors take responsibility for the scientific content.

\FloatBarrier
\bibliographystyle{ieeenat_fullname}
\bibliography{references}

\clearpage
\onecolumn
\raggedbottom
\captionsetup{singlelinecheck=false}
\appendix
\section*{Supplement Overview}

The supplement follows the \trm{} head from logged trajectories to executed actions. Figure~\ref{fig:trm_controller_interface} connects three views of that process: training from temporal pairs, planning with a frozen world model, and auditing the scores on an identical set of candidate actions. The execution sequences below show where choices made by those scores become visible in the environment.

Sections A--C describe the head, the ranking and latent-space diagnostics, and the evaluation protocols. Sections D-- I test alternative explanations and follow the findings across checkpoints, candidate pools, and continuous-control tasks. Sections J--K record the experimental settings and provide the detailed results behind the main-text figures.

\setcounter{figure}{12}
\begin{figure}[H]
\centering
\includegraphics[width=0.98\linewidth]{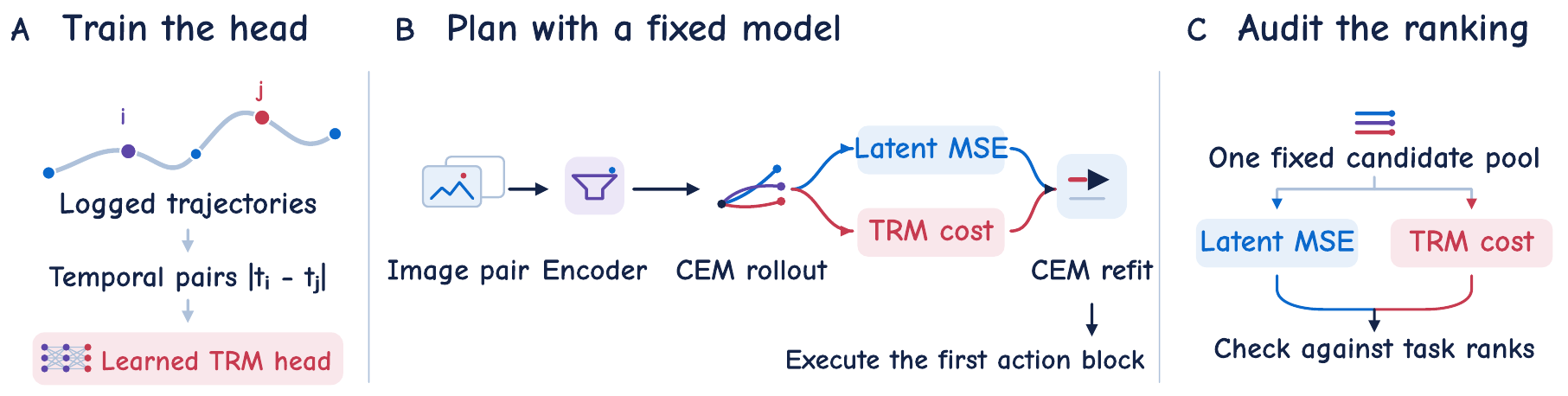}
\caption{Control and analysis interfaces for \trm{}.}
\label{fig:trm_controller_interface}
\end{figure}

\section*{Closed-Loop Execution Sequences}

We sample observations around turns, stalls, and goal arrival. Both methods in a navigation panel start from the same state and goal. Step numbers index executed environment actions; frame numbers index the reference videos.

\setcounter{figure}{13}
\begin{figure}[H]
\centering
\includegraphics[width=0.99\linewidth]{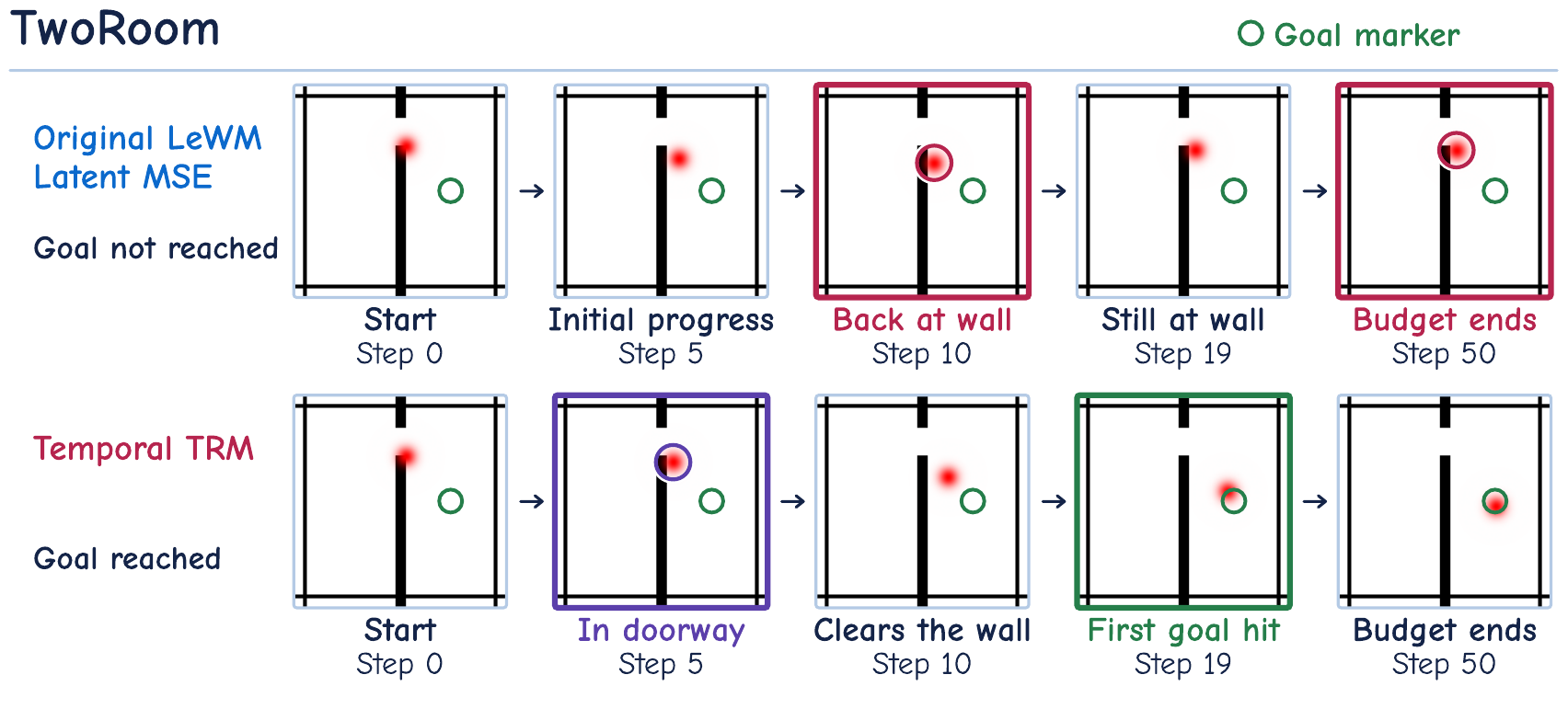}
\caption{TwoRoom: matched execution with raw latent cost and temporal \trm{}.}
\label{fig:execution_14}
\end{figure}

In TwoRoom, both runs begin in the doorway and use 30 \cem{} iterations. The raw-cost controller initially moves toward the goal but returns to the wall by step 10 and ends 59 pixels away. Temporal \trm{} clears the wall, first enters the 16-pixel goal radius at step 19, and ends 5.7 pixels away. This example uses a larger search budget than the default evaluation.

\setcounter{figure}{14}
\begin{figure}[H]
\centering
\includegraphics[width=0.96\linewidth]{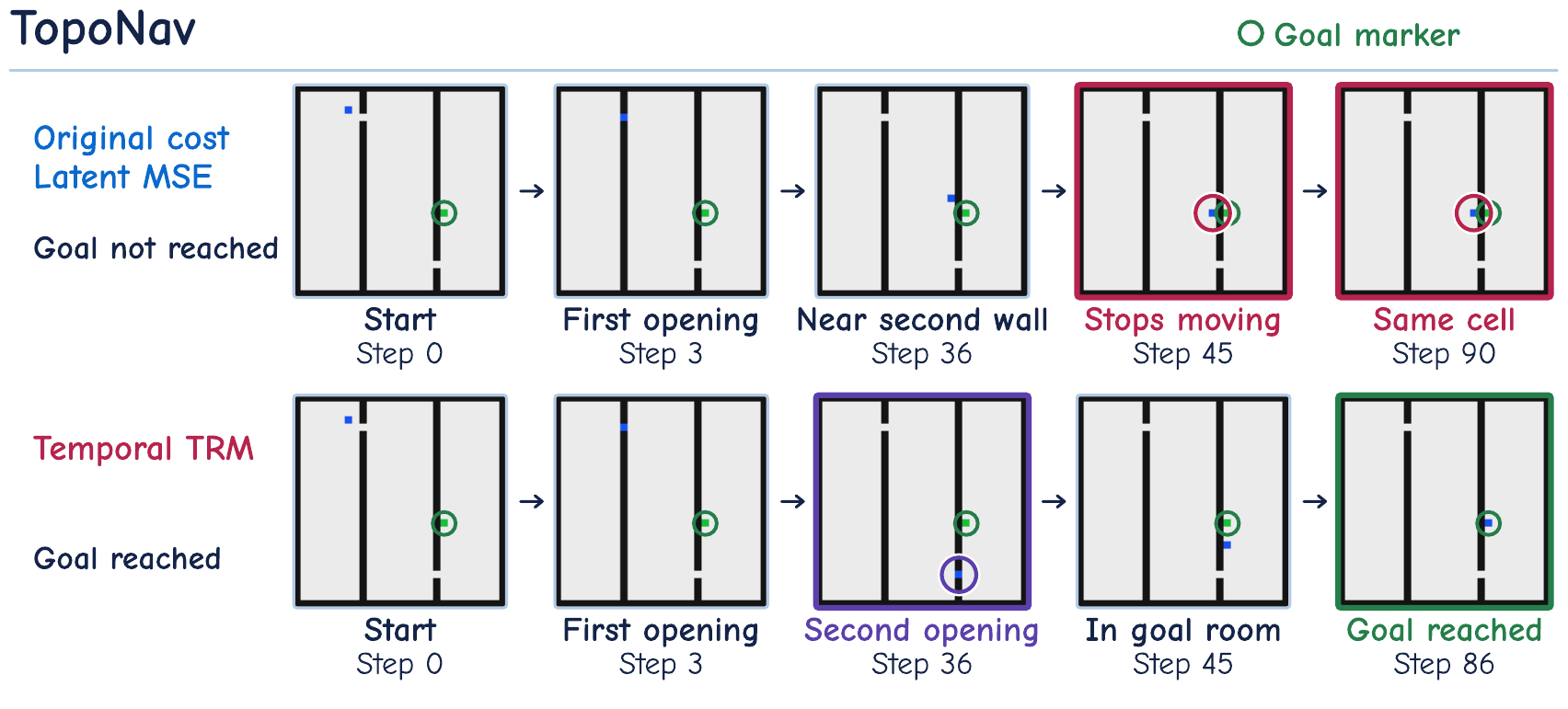}
\caption{TopoNav: matched execution with raw latent cost and temporal \trm{}.}
\label{fig:execution_15}
\end{figure}

The TopoNav panels show matched CPU replays that reproduce the recorded success outcomes, first arrival steps, and final and minimum distances. Raw latent cost moves toward the second wall and then remains in the same cell from step 45 to the 90-step limit, 16 steps from the goal along the shortest path. Temporal \trm{} passes through the upper and lower openings and reaches the goal at step 86 (Section~\ref{app:toponav}).

\setcounter{figure}{15}
\begin{figure}[H]
\centering
\includegraphics[width=0.96\linewidth]{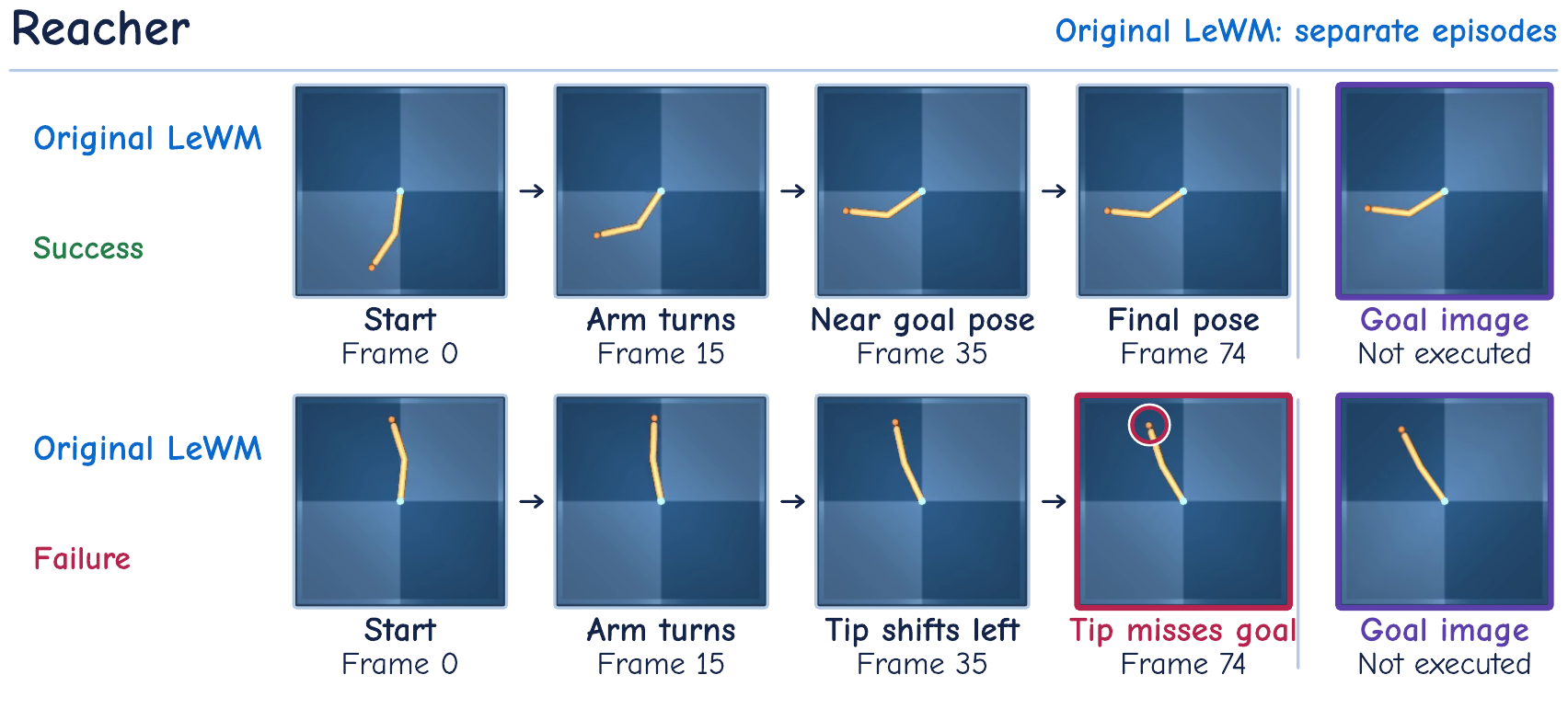}
\caption{Reacher: successful and failed original LeWM executions.}
\label{fig:execution_16}
\end{figure}

In Reacher, the successful run finishes near its goal pose. In the failed run, the highlighted tip remains offset from the corresponding goal image. These are separate original LeWM episodes, each with its own start and goal, not a comparison between methods.

\setcounter{figure}{16}
\begin{figure}[H]
\centering
\includegraphics[width=0.96\linewidth]{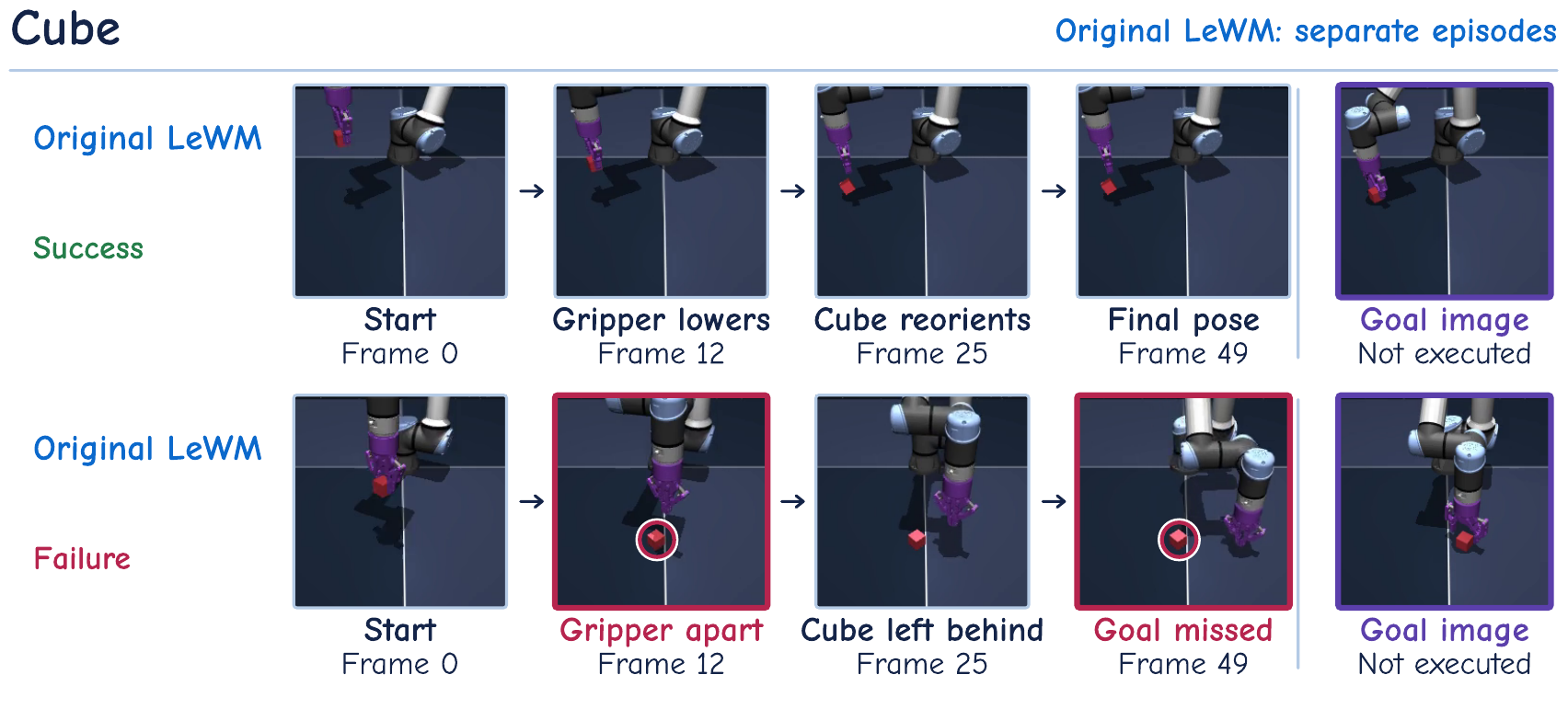}
\caption{Cube: successful and failed original LeWM executions.}
\label{fig:execution_17}
\end{figure}

In the failed Cube episode, the gripper is visibly separated from the cube by frame 12, and the cube remains displaced from its goal at the end. Both examples use original LeWM with separate starts and goals. For the controlled raw-cost versus temporal-head results on Reacher and Cube, see Section~\ref{app:new_continuous}.

The PushT evaluation is not shown as a frame sequence. Section~\ref{app:pusht_details} reports the numerical comparisons, and Figure~\ref{fig:task_execution_schematics} illustrates the contact requirement schematically.

\section{\trm{} Method Details}
\label{app:trm_details}

\setcounter{subsection}{0}
\subsection{Training Objective and Pair Sampling}
\label{app:method_training}

Given a fixed world model encoder $f_\theta$ and latent rollout model $F_\theta$, we train a pairwise scalar head $m_\phi(\z_i,\z_j)$ over encoded latent states. The default feature map is
\begin{equation}
    \psi(\z_i,\z_j)=\left[\z_i,\z_j,\z_i-\z_j,|\z_i-\z_j|\right].
\end{equation}
The head is an \mlp{} with two hidden layers.
\begin{equation}
    m_\phi(\z_i,\z_j)
    =
    \mathrm{\softplus}\!\left(\mathrm{MLP}_\phi(\psi(\z_i,\z_j))\right),
\end{equation}
where $\mathrm{MLP}_\phi$ has two hidden layers with 256 units each and \silu{} nonlinearities. The \softplus{} output enforces nonnegative predicted distance. We train with \smoothlone{} loss on scaled labels.
\begin{equation}
    \min_\phi\; \mathbb{E}_{(i,j)}\;
    \ell_{\mathrm{Huber}}\left(m_\phi(\z_i,\z_j),\; y_{ij}/s\right),
\end{equation}
with scale $s=224$, the \tworoom{} image and coordinate scale used only for loss conditioning.

For temporal sampling in \tworoom{}, we draw an episode $e$ of length $L_e$ and choose $\Delta$ uniformly from $[1,L_e-1]$. The short horizon ablation instead uses $[1,\min(L_e-1,\Delta_{\max})]$. We then choose $t$ uniformly from $[0,L_e-\Delta-1]$ and place rows $(e,t)$ and $(e,t+\Delta)$ in random order. The label is $y_{ij}=\Delta$. Balanced sampling allocates comparable numbers of training pairs to different ranges of temporal separation.

Full episode sampling includes larger temporal separations than the short horizon setting. Table~\ref{tab:appendix_temporal_ablation} changes trajectory coverage, pair count, and maximum temporal separation separately.

\setcounter{table}{4}
\begin{table}[H]
\centering
\caption{TRM head implementation details.}
\label{tab:appendix_trm_hparams}
\begin{tabular}{L{0.30\linewidth}L{0.60\linewidth}}
\toprule
Component & Setting \\
\midrule
Input features & $[\z_i,\z_j,\z_i-\z_j,|\z_i-\z_j|]$ \\
Architecture & \mlp{}, two 256-unit hidden layers \\
Nonlinearity & \silu{} hidden layers, \softplus{} scalar output \\
Training update & \adamw{} \\
Learning rate & $10^{-3}$ \\
Weight decay & $10^{-4}$ \\
Batch size & 1024 pairs \\
Loss & \smoothlone{} on scaled distances \\
Distance scale & 224 \\
Main train pairs & 100,000 \\
Validation pairs & 10,000 independently sampled pairs in the main protocol \\
Checkpoint selection & Lowest validation loss within the fixed epoch budget \\
Shuffled control & Same architecture/data, permuted training labels only \\
\bottomrule
\end{tabular}
\end{table}

\setcounter{subsection}{1}
\subsection{Controller Use}

At evaluation time, \trm{} replaces or augments only the terminal cost. For each \cem{} replan, the controller proposes candidate action sequences $\aseq^{(i)}$. The fixed world model rolls each sequence forward to a predicted terminal latent state $\hat{\z}^{(i)}_{t+H}$. The goal image is encoded once as $\zg$. The replacement \trm{} cost is
\begin{equation}
    c_{\trm}(\aseq^{(i)})=m_\phi\!\left(\hat{\z}^{(i)}_{t+H},\zg\right).
\end{equation}
For the hybrid terminal cost, we standardize the raw latent and learned cost values over the candidate action sequences in the current \cem{} iteration.
\begin{equation}
    c_{\mathrm{hyb}}^{(i)}
    =
    \frac{c_{\mathrm{lat}}^{(i)}-\mu_{\mathrm{lat}}}{\sigma_{\mathrm{lat}}+\epsilon}
    +
    \lambda
    \frac{c_{\trm}^{(i)}-\mu_{\trm}}{\sigma_{\trm}+\epsilon}.
\end{equation}
This standardization uses the current candidate pool. It prevents either component from dominating the terminal cost only because its numerical scale is larger.

We use \emph{head} for the learned function $m_\phi$ and \emph{terminal cost} for the scalar rule that assigns a cost to each candidate action sequence. CEM uses these costs to refit its search distribution and returns the final mean plan. The environment executes the receding-horizon prefix specified by the task protocol.

\setcounter{table}{5}
\begin{table}[H]
\centering
\caption{Complete \tworoom{} controller configuration.}
\label{tab:appendix_controller_config}
\begin{tabular}{L{0.25\linewidth}L{0.63\linewidth}}
\toprule
Controller item & Fixed setting for \tworoom{} \\ 
\midrule
Action space & Two continuous velocity components. The environment clips each component to $[-1,1]$ and multiplies it by the sampled agent speed. The reported evaluation uses the fixed speed recorded in the environment configuration. \\
Planning horizon & 5 model steps with an action block of 5, giving 25 environment actions per candidate action sequence. \\
Replanning & Receding horizon 5. The environment therefore executes all 25 actions from the final \cem{} mean plan before the controller replans. A 50-step evaluation makes two controller calls. \\
\cem{} initialization & Zero mean and unit standard deviation in standardized action coordinates. The first candidate in every iteration is the current mean. \\
\cem{} update & 300 candidates, 10 iterations, 30 elites. The next mean and standard deviation are the unweighted sample mean and standard deviation of the elites. \\
Smoothing and variance floor & No smoothing coefficient and no explicit variance floor. \\
Clipping & \cem{} does not clip actions. The environment applies action clipping after inverse standardization. \\
Success and termination & \tworoom{} success at a Euclidean goal distance below 16 pixels. Evaluation continues for the fixed 50-step reporting budget while retaining the first success step. \\
World model rollout & Frozen encoder and dynamics. The terminal cost uses the final predicted latent after 5 model steps. \\
\sasc{} candidate pool & 300 Gaussian candidate action sequences per episode with candidate seed 123. The candidate pool is generated separately from the final \cem{} iteration, and every terminal cost scores the same candidate action sequences. \\
\bottomrule
\end{tabular}
\end{table}

\setcounter{table}{6}
\begin{table}[H]
\centering
\caption{Pseudocode for training and using TRM.}
\label{tab:appendix_algorithm}
\begin{tabular}{L{0.94\linewidth}}
\toprule
\textbf{Train \trm{}} \\
1. Freeze the world model encoder $f_\theta$ and dynamics $F_\theta$. \\
2. Sample state pairs from logged trajectories using the temporal horizon rule above. \\
3. Encode both observations, $\z_i=f_\theta(o_i)$ and $\z_j=f_\theta(o_j)$. \\
4. Build pair features $[\z_i,\z_j,\z_i-\z_j,|\z_i-\z_j|]$. \\
5. Train $m_\phi$ to predict scaled temporal distance with \smoothlone{} loss. \\
\midrule
\textbf{Use \trm{} in latent \cem{} planning} \\
1. Encode the current observation and goal observation. \\
2. Sample candidate action sequences with the same \cem{} controller. \\
3. Roll each candidate forward through the fixed world model. \\
4. Use $m_\phi$ to compute the terminal cost of each candidate action sequence from its predicted terminal latent state. \\
5. CEM uses the replacement or hybrid costs to update its distribution and returns the final mean plan. \\
6. The environment executes the configured receding-horizon prefix of that plan. \\
\bottomrule
\end{tabular}
\end{table}

\section{Diagnostic Protocols}
\label{app:diagnostics}

\setcounter{subsection}{0}
\subsection{Selection Audit with Shared Candidates}
\label{app:sasc_audit}

For each episode, we fix a pool of candidate action sequences and score each sequence twice. The planner cost $c(i)$ scores the endpoint predicted by the frozen world model against the goal, for example with latent MSE or the learned \trm{} cost. To compute the reference cost $c^\star(i)$, we execute the same actions under the task dynamics and measure the remaining distance to the goal. In \tworoom{}, the reference is the Euclidean distance or the shortest navigable distance, as specified in the corresponding result. TopoNav uses the shortest path length on the grid. \pusht{} combines errors in object position, end-effector position, and object angle at the final simulated state (Section~\ref{app:pusht_details}). The latent and learned planners do not see the reference cost; we compute it separately for the audit. The audit holds the pool fixed and compares top-ranked sequences rather than executing the final \cem{} mean plan.

We compute each metric within an episode and then average over the stated episodes or seeds. In the tables, $\rho$ denotes candidate-order Spearman correlation; subscripts or column labels identify the reference distance. Rank denotes oracle best rank. The audit reports:
\begin{itemize}
\item \textbf{Candidate order Spearman.} Rank the candidates by $c$ and by $c^\star$, give tied candidates their average rank, and compute the Pearson correlation of the two rank lists. Higher correlation means the planner and the reference order the pool more similarly.
\item \textbf{Oracle best rank.} Find the candidate with the smallest reference cost and read its percentile in the planner's ordering. Suppose candidate A has the smallest reference distance in a \tworoom{} pool: it ends 12 pixels from the goal along the shortest navigable route, but ranks 20th by latent MSE. This metric reports A's planner rank as a percentile within the pool. Lower is better.
\item \textbf{Selected diagnostic distance.} Find the candidate ranked first by the planner, then report its reference distance. In the same example, if latent MSE ranks candidate B first and B has 30 pixels left by the same reference, this metric is 30 pixels.
\item \textbf{Top-$k$ diagnostic cost.} Report the reference costs of the $k$ candidates ranked first by the planner.
\end{itemize}
For simulator replay, Appendix~\ref{app:candidate_decomposition} defines separate measures of available progress and selection regret; the \pusht{} comparison in Appendix~\ref{app:pusht_details} evaluates ranking and realized execution separately.

\setcounter{table}{7}
\begin{table}[H]
\centering
\caption{SASC audit on the high distance evaluation set for \lewm{} seed 3072.}
\label{tab:appendix_hard_sasc}
\begin{tabular}{lccc}
\toprule
Terminal cost & $\rho$ Euc. & $\rho$ Geo. & Best rank \\
\midrule
Raw latent MSE & 0.021 & 0.018 & 31.71 \\
Decoded XY & 0.876 & 0.860 & 0.59 \\
\trm{} temporal labels & 0.720 & 0.729 & 3.86 \\
\trm{} shuffled labels & 0.105 & 0.119 & 34.00 \\
\bottomrule
\end{tabular}
\end{table}

\setcounter{subsection}{1}
\subsection{Subspace Projection Audit}
\label{app:subspace_projection}

The linear XY probe reads position from two weighted sums of the latent features. Its row space is the span of those two weight vectors; directions in the orthogonal complement do not affect the probe's output. The probe matrix $W$ defines the orthogonal projection

\begin{equation}
    P_W = W^\top(WW^\top)^\dagger W.
\end{equation}

For each candidate action sequence, let $d=\hat{\z}_{t+H}-\zg$ be its predicted terminal difference. The XY rowspace terminal cost is $\|P_Wd\|_2^2$, and the residual terminal cost is $\|(I-P_W)d\|_2^2$. Table~\ref{tab:appendix_subspace_per_seed} reports the 50-action distance-controlled intervention by seed.

\setcounter{table}{8}
\begin{table}[H]
\centering
\caption{Per seed subspace intervention on the distance controlled set with a budget of 50 actions.}
\label{tab:appendix_subspace_per_seed}
\begin{tabular}{rrrrrr}
\toprule
Seed & Latent MSE & Decoded XY & XY rowspace & Residual only & Rowspace MSE share \\
\midrule
3072 & 2.5 & 95.0 & 97.5 & 2.5 & 0.006 \\
3073 & 2.5 & 95.0 & 92.5 & 2.5 & 0.0069 \\
3074 & 0.0 & 92.5 & 82.5 & 0.0 & 0.005 \\
\bottomrule
\end{tabular}
\end{table}

Table~\ref{tab:episode_disjoint_xy} repeats the intervention with episode-disjoint probes.

\setcounter{subsection}{2}
\subsection{Direct Attribution of the \trm{} Terminal Cost}
\label{app:trm_rowspace_attribution}

For each candidate action sequence, the frozen world model predicts a terminal latent state $\hat{\z}_{t+H}$. We write its difference from the goal latent as $d=\hat{\z}_{t+H}-\zg$. We then construct two diagnostic terminal latent states. The first is $\zg+P_Wd$, which retains only the part of the candidate difference in the XY rowspace. The second is $\zg+(I-P_W)d$, which retains only the residual part. The same trained head evaluates each constructed terminal latent state against $\zg$ and assigns the resulting cost to the same candidate action sequence. The candidate action sequences, predicted terminal latent states, and head parameters remain fixed. We measure the fraction of the squared cost-gradient norm that lies in the XY directions by differentiating the original \trm{} terminal cost with respect to $d$:

\begin{equation}
 r_W=\frac{\|P_W\nabla_d m_\phi(\zg+d,\zg)\|_2^2}
 {\|\nabla_d m_\phi(\zg+d,\zg)\|_2^2}.
\end{equation}

The audit uses heads trained with episode exclusion. Every head scores the same 300 candidates per episode, using Euclidean endpoint distance as the reference in Table~\ref{tab:appendix_trm_rowspace_attribution} and Figure~\ref{fig:appendix_trm_attribution}. An isotropic gradient places $2/192\approx1.04\%$ of its squared norm in the two XY directions on average. The observed fractions are 3.4 to 4.5 times this baseline.

\setcounter{table}{9}
\begin{table}[H]
\centering
\caption{Direct TRM attribution on the high distance candidate pools.}
\label{tab:appendix_trm_rowspace_attribution}
\begin{tabular}{rrrrrrrr}
\toprule
Seed & Full $\rho$ & Row $\rho$ & Resid. $\rho$ & Full rank & Row rank & Resid. rank & $r_W$ \\
\midrule
3072 & 0.615 & 0.569 & 0.587 & 9.44 & 14.90 & 9.14 & 3.49\% \\
3073 & 0.539 & 0.228 & 0.511 & 5.23 & 43.04 & 6.15 & 4.64\% \\
3074 & 0.583 & 0.325 & 0.559 & 5.59 & 32.76 & 5.65 & 3.70\% \\
\bottomrule
\end{tabular}
\end{table}

\setcounter{figure}{17}
\begin{figure}[H]
\centering
\includegraphics[width=1.00\linewidth]{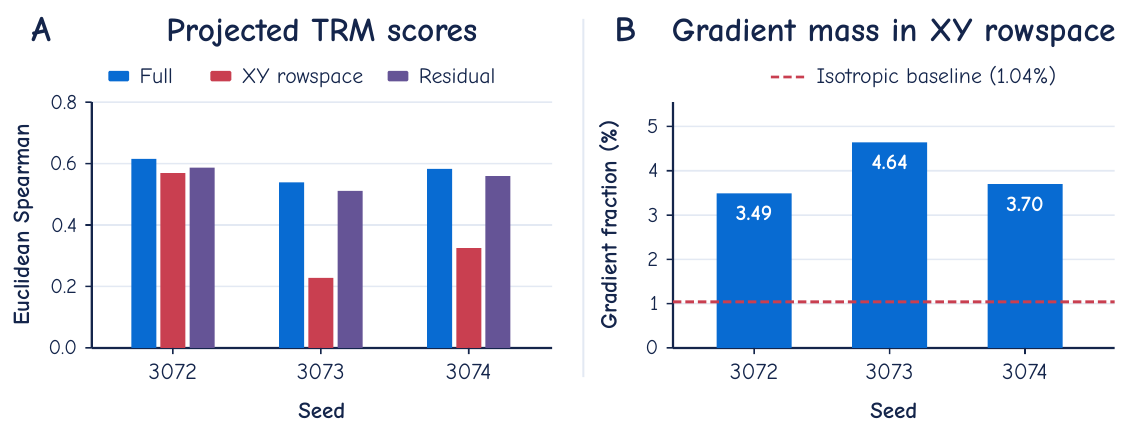}
\caption{\trm{} attribution: (A) candidate ranking by input projection; (B) XY share of the squared cost gradient.}
\label{fig:appendix_trm_attribution}
\end{figure}

The projected-input interventions complement the gradient test: residual features retain useful ordering, and XY-only ordering is weaker for two seeds. The head is sensitive to position without depending exclusively on the XY probe subspace.

\setcounter{subsection}{3}
\subsection{Swap Consistency and Feature Groups}
\label{app:trm_feature_intervention}

The time gap is the same in either input order, but the head can return different scores when we swap its inputs. On the high distance candidate pools, the mean absolute difference between $m_\phi(\z_i,\z_j)$ and $m_\phi(\z_j,\z_i)$ is 4.92, 6.13, and 5.35 temporal steps for seeds 3072, 3073, and 3074. These values are 6.4\%, 7.7\%, and 6.9\% of the mean terminal cost magnitude. We next use the same trained head to score the same candidate action sequences while changing only its input features. For each row in Table~\ref{tab:appendix_feature_intervention}, we retain the named feature blocks and set every other block to zero before the head assigns a cost. The controller then ranks the unchanged candidate action sequences by these costs. We do not retrain the head or select a new checkpoint for this intervention.

To test whether combining the two input orders improves ranking, we also average their scores:
\begin{equation}
m_{\mathrm{sym}}(\z_i,\z_j)=\tfrac12[m_\phi(\z_i,\z_j)+m_\phi(\z_j,\z_i)].
\end{equation}

We use the Euclidean reference from Section~\ref{app:trm_rowspace_attribution}. Absolute differences alone do not preserve the candidate ordering. Using both signed and absolute differences helps seed 3073 but not the other two seeds. Full features retain useful ordering across seeds, and averaging the two orders improves the oracle best rank for each seed.

\setcounter{table}{10}
\begin{table}[H]
\centering
\caption{Feature group and input-order interventions.}
\label{tab:appendix_feature_intervention}
\begin{tabular}{lrrrrrr}
\toprule
Input to fitted head & \multicolumn{2}{c}{Seed 3072} & \multicolumn{2}{c}{Seed 3073} & \multicolumn{2}{c}{Seed 3074} \\
 & $\rho$ & Rank & $\rho$ & Rank & $\rho$ & Rank \\
\midrule
$|z_i-z_j|$ only & 0.006 & 27.47 & $-0.028$ & 36.80 & 0.074 & 45.93 \\
$[z_i-z_j,|z_i-z_j|]$ only & 0.037 & 42.45 & 0.555 & 8.27 & 0.086 & 19.07 \\
Full ordered features & 0.615 & 9.44 & 0.539 & 5.23 & 0.583 & 5.59 \\
Full features with swap symmetrization & 0.738 & 3.40 & 0.592 & 3.33 & 0.686 & 3.24 \\
\bottomrule
\end{tabular}
\end{table}

\setcounter{subsection}{4}
\subsection{Temporal Labels and Geodesic Distance}
\label{app:temporal_geodesic}

We use the same episode exclusion as in the high-distance audit. Across logged pairs, larger time gaps generally correspond to longer navigable distances (Table~\ref{tab:appendix_temporal_geodesic}).

\setcounter{table}{11}
\begin{table}[H]
\centering
\caption{Correlation between logged time gaps and \tworoom{} geodesic distance.}
\label{tab:appendix_temporal_geodesic}
\begin{tabular}{lrr}
\toprule
Pair subset & Pairs & Spearman \\
\midrule
All & 100,000 & 0.883 \\
Same room & 70,976 & 0.854 \\
Doorway required & 29,024 & 0.779 \\
\bottomrule
\end{tabular}
\end{table}

\section{Experimental Details}
\label{app:experimental_details}

\setcounter{subsection}{0}
\subsection{Data Appendix and Public Sources}
\label{app:data_appendix}

The experiments use three data sources. For \tworoom{}, our main navigation setting, we use the expanded public dataset (632,749 observations from 10,000 episodes), not the smaller 493-observation default dataset~\citep{maes2026stableworldmodel,maes2026leworldmodel}. The frozen LeWM models use 192-dimensional latents, 512 projections, and training seeds 3072, 3073, and 3074. We use the same models to encode logged temporal pairs, construct the evaluation sets, and generate the \sasc{} candidate pools.

We construct TopoNav as a grid navigation diagnostic with known shortest-path distances. Section~\ref{app:toponav} specifies the grid, rendering, state representation, logged pair construction, and controller.

For \pusht{}, we use public expert demonstrations and the frozen LeWM model from the stable world model release~\citep{chi2023diffusionpolicy,maes2026stableworldmodel}. We evaluate goals sampled 25, 50, and 75 steps ahead and study whether ranking improvements translate into better execution.

Figure~\ref{fig:appendix_task_suite} contrasts two kinds of navigation with contact manipulation. In \tworoom{}, the planner must prefer a route through the doorway to a straight line blocked by the wall. TopoNav uses a grid with exact graph distances. In \pusht{}, moving the object requires contact, so an improvement in candidate ordering need not improve closed-loop execution.

\setcounter{figure}{19}
\begin{figure}[H]
\centering
\includegraphics[width=0.98\linewidth]{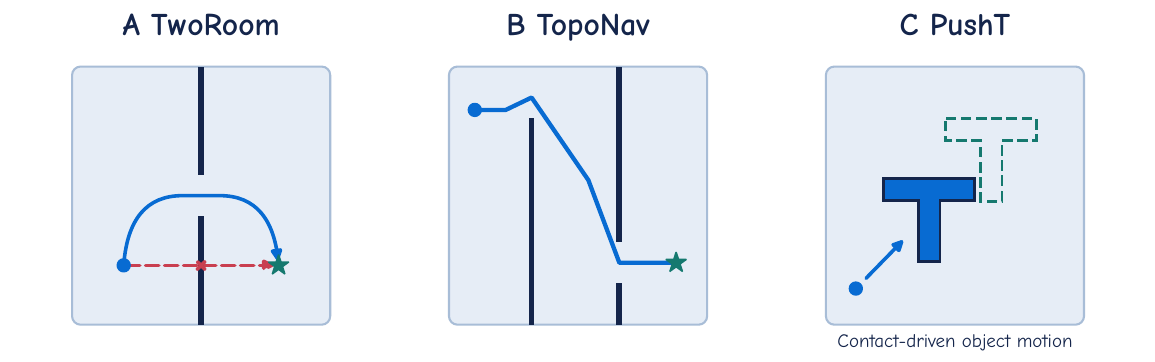}
\caption{Task suite geometry.}
\label{fig:appendix_task_suite}
\end{figure}

The \tworoom{} and \pusht{} observations and frozen models come from the cited public releases. No experiment requires a private dataset.

The new data products introduced by this paper are the TopoNav trajectories, fixed evaluation sets, logged pair indices and labels, \sasc{} candidate audit records, and aggregate tables derived from those records. TopoNav trajectories are generated from the grid and transition rules in Section~\ref{app:toponav}. The other new records contain indices, model outputs, controller costs, or task metrics derived from the cited public sources. No human participants, personal data, or manual annotation are used. The \tworoom{} and \pusht{} observations and checkpoints retain the licenses of their source releases, and we do not relicense those upstream assets.

We do not retrain the \lewm{} or \pldm{} world models. Reproducing our comparisons begins from the cited frozen checkpoints. The source releases define the model architectures and their original training procedures. This supplement specifies the checkpoint family, latent interface, controller settings, and added \trm{} head that affect the reported experiments.

\setcounter{subsection}{1}
\subsection{\tworoom{} Evaluation Sets and Failure Taxonomy}
\label{app:tworoom_sets}

The balanced evaluation set contains 40 episodes, with 20 goals in the same room and 20 goals across the wall. The distance controlled evaluation set uses the same counts and draws both groups from the same Euclidean distance range. The high distance evaluation set contains 100 episodes, with 50 goals in the same room and 50 goals across the wall. We subsequently refer to them as the balanced, distance controlled, and high distance sets. Among the 50 goals across the wall, 47 require the doorway.

B50/B150 and D50/D150 denote the balanced and distance-controlled sets with 50/150 executed actions, respectively.

Evaluation set construction computes Euclidean and approximate geodesic distance from the recorded start and goal coordinates. Distance controlled evaluation sets draw equal same room and cross wall counts from the same Euclidean distance range. The high distance set uses geodesic distances from 120 to 205 pixels and requires a margin of 5 to 30 steps relative to the action budget of 50. A pair is doorway required when the endpoints lie on opposite wall sides and the straight segment intersects the solid wall outside every door interval.

We use the following failure taxonomy. Success occurs when the environment reaches a goal distance below 16 pixels. Wrong room means the final agent location remains on the wrong side of the wall. Stuck at wall means the final position lies within one speed unit of the wall and outside a door interval. Same room not precise and crossed door not precise are near miss categories used in episode reports.

\setcounter{subsection}{3}
\subsection{Linear XY Probe Details}
\label{app:xy_probe}

The XY probe uses 20,000 rows sampled without replacement from the public cache. We randomly assign 16,000 sampled rows to training and 4,000 to testing. Ridge regression uses $\alpha$ = 1. This row-level test measures position decodability; the separate head audit in Appendix~\ref{app:tworoom_nooverlap} excludes entire evaluation episodes.

The probe error in Table~\ref{tab:appendix_xy_probe} is measured in pixels.

\setcounter{table}{13}
\begin{table}[H]
\centering
\caption{Linear XY probe results.}
\label{tab:appendix_xy_probe}
\begin{tabular}{rrrr}
\toprule
Seed & Test RMSE & Test $R^2$ & Train/Test rows \\
\midrule
3072 & 1.776 & 0.99818 & 16k/4k \\
3073 & 1.870 & 0.99802 & 16k/4k \\
3074 & 1.855 & 0.99801 & 16k/4k \\
\bottomrule
\end{tabular}
\end{table}

\textbf{Episode-Disjoint Validation of the XY Subspace.} To test whether the XY rowspace advantage depends on trajectory overlap, we refit the probe without any of the 193 trajectories that supply start or goal observations for the 100-case set. For the 40-case set, a separate fit excludes the endpoint source trajectories of both sets (267 in total). On the held-out trajectories for the 100-case set, $R^2$ remains 0.9975--0.9979 across three model seeds.

We repeat the intervention in Section~\ref{app:subspace_projection} with the refitted probes, holding the start--goal cases, world model, and controller settings fixed within each comparison. Only the terminal cost changes.

\setcounter{table}{13}
\begingroup
\renewcommand{\thetable}{14a}
\renewcommand{\theHtable}{supp.14a}
\begin{table}[H]
\centering
\caption{Closed-loop success with episode-disjoint XY probes.}
\label{tab:episode_disjoint_xy}
\begin{tabular}{lcc}
\toprule
Terminal cost & 40-case set & 100-case set \\
\midrule
Raw latent MSE & 3.3\% (4/120) & 6.3\% (19/300) \\
XY rowspace & \textbf{89.2\%} (107/120) & \textbf{92.7\%} (278/300) \\
Orthogonal residual & 1.7\% (2/120) & 5.7\% (17/300) \\
\bottomrule
\end{tabular}
\end{table}
\endgroup

Counts pool the same cases across three model seeds. The refitted XY rowspace substantially outperforms both raw latent MSE and its orthogonal residual on both sets. The control advantage therefore does not require the probe to have seen the source trajectories of the evaluation endpoints.

\setcounter{subsection}{4}
\subsection{Raw Latent Controller Budget}
\label{app:raw_latent_budget}

On the 40 distance-controlled episodes with a 50-action execution budget, raising the search budget for \lewm{} seed 3072 leaves aggregate raw-cost success at 2.5\%. The default budget succeeds on 5.0\% of same-room cases and none of the cross-wall cases; with the larger budget those rates are 0.0\% and 5.0\%, respectively.

\setcounter{table}{14}
\begin{table}[H]
\centering
\caption{Raw latent CEM search-budget comparison.}
\label{tab:appendix_solver_budget}
\begin{tabular}{lrrrr}
\toprule
Setting & Samples & Iter. & Elite & Success \\
\midrule
Default & 300 & 10 & 30 & 2.5 \\
Larger budget & 1000 & 20 & 100 & 2.5 \\
\bottomrule
\end{tabular}
\end{table}

\setcounter{subsection}{7}
\subsection{Compute Infrastructure}
\label{app:compute}

We use NVIDIA RTX 6000 Ada GPUs for our experiments.

\setcounter{subsection}{8}
\subsection{Reporting Units, Repeats, and Termination}
\label{app:reporting_protocol}

Closed-loop success is measured per episode. Across three seeds, we report the mean and sample standard deviation of seed-level success rates. The high-distance \sasc{} intervals use 100,000 bootstrap resamples paired by episode, retaining that episode's full candidate pool. For TopoNav success, the bootstrap resamples head seeds and the 120 paired episodes within each selected seed. Table~\ref{tab:appendix_statistical_intervals} reports the 95\% intervals.

\setcounter{table}{17}
\begin{table}[H]
\centering
\caption{Means and 95\% paired bootstrap intervals.}
\label{tab:appendix_statistical_intervals}
\begin{tabular}{L{0.25\linewidth}L{0.20\linewidth}L{0.20\linewidth}L{0.22\linewidth}}
\toprule
Audit quantity & Raw latent MSE & Temporal \trm{} & Paired difference, \trm{} minus raw \\
\midrule
High distance evaluation \sasc{} $\rho_{\mathrm{geo}}$ & 0.018 $[-0.053,0.088]$ & 0.729 $[0.695,0.759]$ & 0.711 $[0.644,0.779]$ \\
High distance evaluation oracle best rank, \% & 31.71 $[26.49,37.11]$ & 3.86 $[2.37,5.62]$ & $-27.85$ $[-33.63,-22.16]$ \\
High distance evaluation selected geodesic & 168.81 $[163.02,174.64]$ & 134.79 $[130.41,139.40]$ & $-34.02$ $[-39.40,-28.69]$ \\
TopoNav success, \% & 15.0 $[11.39,18.89]$ & 87.5 $[82.50,92.22]$ & 72.5 $[66.39,78.33]$ \\
\bottomrule
\end{tabular}
\end{table}

Under the 50-action \tworoom{} protocol, an episode records the first goal hit and continues to the budget; the final state determines the failure category when there is no hit. TopoNav stops at the exact goal cell or after 90 actions. For \pusht{}, threshold success counts a hit at any step, while final-state success applies the task thresholds to the last state (Section~\ref{app:pusht_details}).

\setcounter{subsection}{9}
\subsection{\tworoom{} Audit after Excluding Evaluation Episodes}
\label{app:tworoom_nooverlap}

Table~\ref{tab:appendix_nooverlap_aggregate} tests whether episode overlap between training and evaluation explains the high distance set result. Before sampling training or validation pairs, we exclude all 193 source episodes that contain any of the 200 evaluation start and goal rows. The encoder, dynamics, high distance set, controller, \cem{} budget, and candidate pools remain fixed. Only the episodes available for temporal pair training change.

\setcounter{table}{19}
\begin{table}[H]
\centering
\caption{High distance evaluation audit with episode exclusion.}
\label{tab:appendix_nooverlap_aggregate}
\begin{tabular}{lrrrr}
\toprule
Terminal cost & Succ. & $\rho_{\mathrm{geo}}$ & Rank & Dist. \\
\midrule
Raw latent MSE & $7.0{\pm}8.2$ & 0.007 & 38.95 & 139.72 \\
\trm{}, episodes excluded & $96.7{\pm}2.1$ & 0.577 & 5.53 & 10.68 \\
Shuffled, episodes excluded & $0.0{\pm}0.0$ & 0.003 & 56.63 & 150.55 \\
\bottomrule
\end{tabular}
\end{table}

The sampled training and validation pairs contain no rows from the excluded episodes.

\setcounter{subsection}{10}
\subsection{TopoNav Diagnostic}
\label{app:toponav}

TopoNav uses zero indexed grid coordinates (x, y) on a 29 $\times$ 29 map. The outer boundary is solid. Two vertical walls occupy x = 9 and x = 19 for interior rows, with single cell openings at (9, 4) and (19, 24). The map has 677 free states. The action set is stay, up, down, left, and right. A move into a wall leaves the state unchanged. Each cell is rendered as a 5 $\times$ 5 RGB block, producing a 145 $\times$ 145 observation. Free space has intensity (0.92, 0.92, 0.92), walls use (0.08, 0.08, 0.08), the goal uses (0.10, 0.75, 0.22), and the agent uses (0.10, 0.32, 0.95).

Logged trajectories follow graph shortest paths between uniformly sampled free cells. The training run uses 2,500 trajectories, 60,000 training pairs, and 12,000 validation pairs. Of the sampled pairs, 85\% use trajectory endpoints and 15\% use two random positions from the same trajectory. Pair order is randomized, and the label is the absolute step separation. The 120 evaluation pairs have graph distance of at least 18, a graph to Euclidean distance ratio of at least 1.45, and Manhattan separation of at least 12. Their exact unordered start and goal pairs are excluded from both training and validation.

The reported run uses base seed 608 and head seeds 3072, 3073, and 3074. Each head has two hidden layers of width 128 and trains for 24 epochs. We retain the checkpoint with the lowest validation loss. The controller uses horizon 12, 256 candidate action sequences, 5 \cem{} iterations, 32 retained candidate action sequences, categorical prior 0.2, and momentum 0.2.

The controller selects the candidate action sequence with the lowest cost. The environment executes its first action, after which the controller replans. The episode stops when the agent reaches the exact goal cell or after 90 executed actions.

TopoNav success requires reaching the exact goal cell, and final distance measures the graph distance left after the last executed action. Across head seeds 3072, 3073, and 3074, temporal \trm{} reaches 87.5\% closed-loop success, versus 15.0\% for raw latent MSE and 1.1\% for shuffled \trm{}.

\setcounter{table}{21}
\begin{table}[H]
\centering
\caption{TopoNav results across head seeds.}
\label{tab:appendix_toponav}
\begin{tabular}{lrrrr}
\toprule
Terminal cost & Succ. & $\rho_{\mathrm{geo}}$ & Rank & Final geo. \\
\midrule
Raw latent MSE & 15.0 & 0.496 & 45.03 & 36.18 \\
Temporal \trm{} & 87.5 & 0.848 & 1.29 & 3.01 \\
Shuffled \trm{} & 1.1 & 0.060 & 49.19 & 42.35 \\
Oracle geodesic & 95.6 & 1.000 & 0.00 & 0.08 \\
\bottomrule
\end{tabular}
\end{table}

\setcounter{subsection}{11}
\subsection{Task Execution Visualizations}
\label{app:task_execution}

Figure~\ref{fig:task_execution_schematics} shows the route constraint in \tworoom{} and the contact requirement in \pusht{}. In \tworoom{}, the red straight line is blocked and the blue route passes through the doorway. In \pusht{}, the arm must make contact to move the T-shaped object. All quantitative results come from the fixed evaluation sets and \sasc{} diagnostics.

On the long-distance set, overall success, wrong-room endings, and wall stalls use all 100 cases as their denominator. The same-room and cross-wall success rates each use their own 50 cases.

For each of the three \lewm{} seeds, the cost baselines share the evaluation set, controller and candidate pools. For these baselines and the changed-layout comparison, ranking uses the geodesic reference in Section~\ref{app:sasc_audit}; Dist. is the final task distance after closed-loop execution.

\setcounter{figure}{21}
\begin{figure}[H]
\centering
\includegraphics[width=0.72\linewidth]{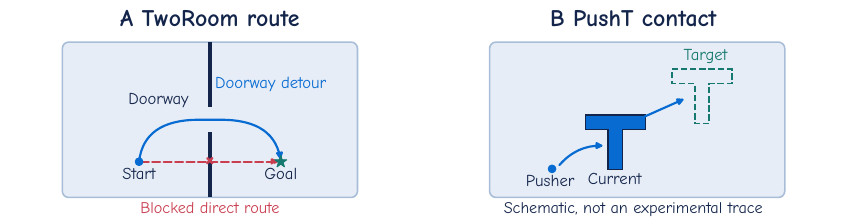}
\caption{Task execution schematics for TwoRoom and PushT.}
\label{fig:task_execution_schematics}
\end{figure}

\setcounter{table}{22}
\begin{table}[H]
\centering
\caption{Closed-loop success on the balanced and distance-controlled \tworoom{} sets.}
\label{tab:appendix_fullcache_per_seed}
\begin{tabular}{lrrrrr}
\toprule
Checkpoint & Seed & B50 & B150 & D50 & D150 \\
\midrule
\lewm{} np512 & 3072 & 55.0 & 60.0 & 2.5 & 17.5 \\
\lewm{} np512 & 3073 & 57.5 & 60.0 & 2.5 & 7.5 \\
\lewm{} np512 & 3074 & 52.5 & 57.5 & 0.0 & 5.0 \\
\pldm{} full e10 & 3072 & 65.0 & 75.0 & 15.0 & 27.5 \\
\pldm{} full e10 & 3073 & 67.5 & 77.5 & 17.5 & 27.5 \\
\pldm{} full e10 & 3074 & 67.5 & 77.5 & 25.0 & 32.5 \\
\bottomrule
\end{tabular}
\end{table}

\setcounter{table}{23}
\begin{table}[H]
\centering
\caption{High distance \tworoom{} outcomes by checkpoint and terminal cost.}
\label{tab:appendix_hardn100_per_seed}
\begin{tabular}{rlrrrrr}
\toprule
Seed & Terminal cost & Success & Same & Cross & Wrong room & Stuck wall \\
\midrule
3072 & \lewm{} latent MSE & 16.0 & 26.0 & 6.0 & 44.0 & 14.0 \\
3073 & \lewm{} latent MSE & 5.0 & 4.0 & 6.0 & 45.0 & 12.0 \\
3074 & \lewm{} latent MSE & 0.0 & 0.0 & 0.0 & 47.0 & 5.0 \\
3072 & \lewm{} \trm{} & 99.0 & 100.0 & 98.0 & 0.0 & 1.0 \\
3073 & \lewm{} \trm{} & 99.0 & 100.0 & 98.0 & 0.0 & 0.0 \\
3074 & \lewm{} \trm{} & 93.0 & 86.0 & 100.0 & 0.0 & 0.0 \\
3072 & \lewm{} shuffled \trm{} & 0.0 & 0.0 & 0.0 & 39.0 & 15.0 \\
3073 & \lewm{} shuffled \trm{} & 0.0 & 0.0 & 0.0 & 49.0 & 11.0 \\
3074 & \lewm{} shuffled \trm{} & 0.0 & 0.0 & 0.0 & 50.0 & 0.0 \\
3072 & \pldm{} latent MSE & 29.0 & 4.0 & 54.0 & 17.0 & 2.0 \\
3072 & \pldm{} \trm{} & 74.0 & 58.0 & 90.0 & 1.0 & 2.0 \\
3072 & \pldm{} shuffled \trm{} & 0.0 & 0.0 & 0.0 & 50.0 & 0.0 \\
3073 & \pldm{} latent MSE & 34.0 & 0.0 & 68.0 & 12.0 & 15.0 \\
3073 & \pldm{} \trm{} & 85.0 & 90.0 & 80.0 & 8.0 & 2.0 \\
3073 & \pldm{} shuffled \trm{} & 0.0 & 0.0 & 0.0 & 50.0 & 0.0 \\
3074 & \pldm{} latent MSE & 35.0 & 8.0 & 62.0 & 9.0 & 25.0 \\
3074 & \pldm{} \trm{} & 93.0 & 90.0 & 96.0 & 0.0 & 2.0 \\
3074 & \pldm{} shuffled \trm{} & 0.0 & 0.0 & 0.0 & 50.0 & 8.0 \\
\bottomrule
\end{tabular}
\end{table}

\setcounter{table}{24}
\begin{table}[H]
\centering
\caption{High distance evaluation terminal cost baselines.}
\label{tab:appendix_hardn100_metric_baselines}
\begin{tabular}{llrrrr}
\toprule
Terminal cost & Supervision & Succ. & $\rho_{\mathrm{geo}}$ & Rank & Dist. \\
\midrule
Raw latent MSE & None & 7.0 & 0.007 & 38.95 & 139.72 \\
Standardized L2 & Latent statistics & 3.3 & 0.007 & 40.34 & 147.59 \\
Linear abs diff & Geodesic & 15.3 & 0.354 & 17.81 & 88.75 \\
Diag. Mahalanobis & Geodesic & 21.3 & 0.279 & 18.93 & 99.37 \\
Decoded XY geodesic & XY probe & 93.3 & 0.857 & 0.56 & 10.90 \\
Temporal \trm{} & Logged temporal & 97.0 & 0.678 & 3.79 & 7.85 \\
Pairwise \mlp{} \trm{} & Geodesic & 99.3 & 0.908 & 0.45 & 7.61 \\
Ranking loss \trm{} & Geodesic & 95.7 & 0.839 & 0.70 & 9.22 \\
Shuffled temporal & Shuffled temporal & 0.0 & 0.053 & 44.10 & 138.86 \\
Oracle geodesic & Oracle upper bound & 100.0 & 1.000 & 0.00 & 5.74 \\
\bottomrule
\end{tabular}
\end{table}

\setcounter{subsection}{12}
\subsection{Coverage, Baseline, and Layout Checks}
\label{app:coverage_layout_checks}

The coverage experiment trains a geodesic pair head for one checkpoint seed. It is not the default temporal \trm{} experiment. All rows use checkpoint seed 3072, the same high distance set, the same controller budget, and the same \sasc{} candidate pool. The limited doorway coverage setting samples 25,000 states that are at least 95 pixels from the nearest door. The balanced doorway coverage setting samples 12,500 states within 70 pixels of a door and 12,500 states farther away. Its 50,000 training pairs contain equal numbers of pairs with and without an endpoint near a door, and 50\% of the pairs cross the wall. Mixed sampling uses 50\% cross wall pairs without constraining distance to a door. Validation uses 10,000 pairs sampled independently from the training pairs.

Under balanced doorway coverage, mixed sampling or random sampling, the oracle-best candidate ranks near the top and control succeeds in every episode. With limited doorway coverage, the head records no successes. On shifted-doorway layouts the geodesic head transfers partly, while none of the tested terminal costs ranks candidates reliably on the rotated-wall layout.

\setcounter{figure}{22}
\begin{figure}[H]
\centering
\includegraphics[width=0.90\linewidth]{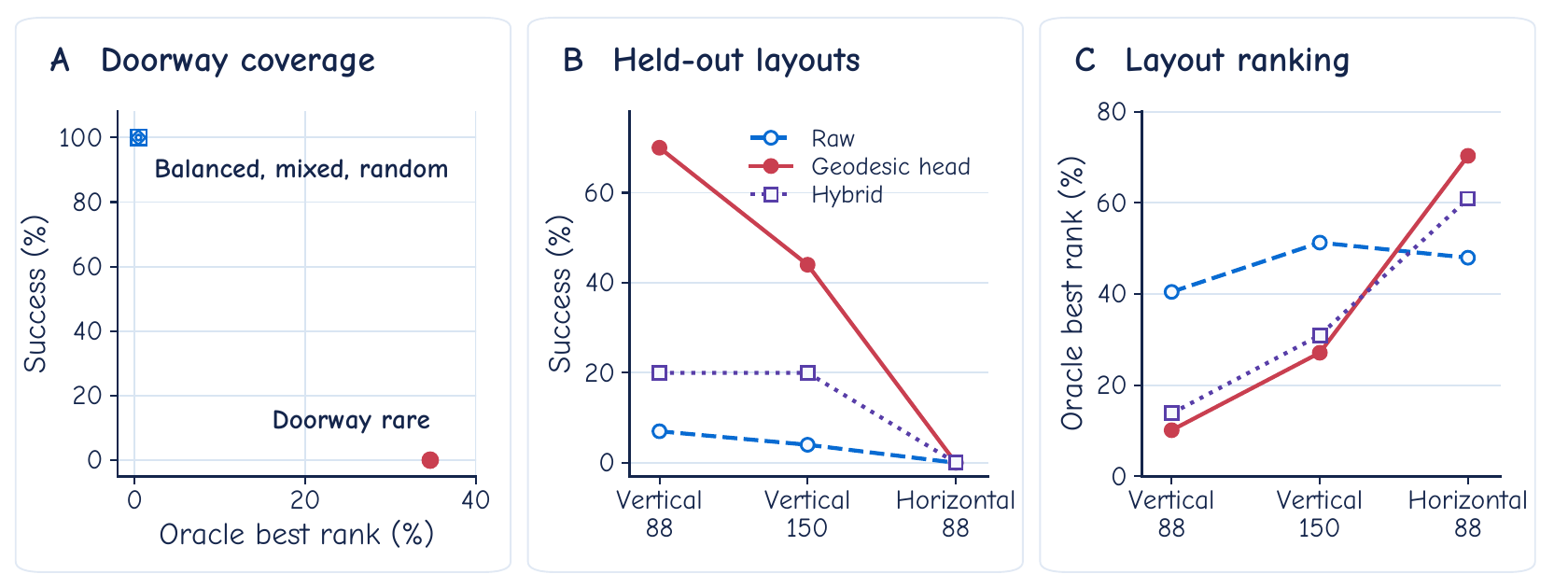}
\caption{Doorway coverage and layout transfer diagnostics.}
\label{fig:appendix_coverage_layout_evidence}
\end{figure}

\setcounter{table}{25}
\begin{table}[H]
\centering
\caption{Doorway coverage diagnostic with geodesic labels.}
\label{tab:appendix_coverage_layout}
\begin{tabular}{llrrrrr}
\toprule
Coverage & Rows & Train pairs & Cross & Near door & Succ. & Rank \\
\midrule
Limited doorway coverage & 25k & 50k & 49.8\% & 0\% & 0 & 34.62 \\
Balanced doorway coverage & 25k & 50k & 50.0\% & 50\% & 100 & 0.49 \\
Mixed & 25k & 50k & 50.0\% & Unconstrained & 100 & 0.43 \\
Random & 30k & 100k & 50.0\% & Unconstrained & 100 & 0.45 \\
\bottomrule
\end{tabular}
\end{table}

\textbf{Coverage under Temporal Supervision.} We train two \trm{} heads on elapsed-step labels, varying how often training pairs include a near-door endpoint. The two training sets match in size, temporal-separation distribution, source trajectories, and number of cross-wall pairs. With the same frozen world model, evaluation cases, and controller budget, enriching near-door endpoint coverage raises success from 7.0\% to 60.0\% (Table~\ref{tab:temporal_coverage}), extending the coverage result to temporal supervision.

\setcounter{table}{25}
\begingroup
\renewcommand{\thetable}{26a}
\renewcommand{\theHtable}{supp.26a}
\begin{table}[H]
\centering
\caption{Success under temporal supervision on the same 100 high-distance TwoRoom cases, using one checkpoint seed.}
\label{tab:temporal_coverage}
\begin{tabular}{lccc}
\toprule
Endpoint coverage & Overall (\%) & Same room (\%) & Cross wall (\%) \\
\midrule
Far-door & 7.0 & 2.0 & 12.0 \\
Near-door & 60.0 & 94.0 & 26.0 \\
\bottomrule
\end{tabular}
\end{table}
\endgroup

The layout study uses checkpoint seed 3072 and 100 episodes per layout at a fixed controller budget. The geodesic pair head is trained on the original vertical wall, with no selection on changed-layout outcomes. Among the raw, geodesic, and hybrid scores in Table~\ref{tab:appendix_layout_full}, the geodesic head has the highest success on both shifted-doorway layouts; all three scores yield 0\% after wall rotation. This is a geodesic-head transfer test, not a test of temporal \trm{} on an unseen layout.

\setcounter{table}{26}
\begin{table}[H]
\centering
\caption{Closed-loop and ranking results on changed layouts.}
\label{tab:appendix_layout_full}
\begin{tabular}{@{}llrrrr@{}}
\toprule
Layout & Cost & Succ. & $\rho_{\mathrm{geo}}$ & Rank & Dist. \\
\midrule
Door 88 & Raw & 7 & 0.130 & 40.46 & 128.04 \\
Door 88 & Geodesic & 70 & 0.777 & 10.11 & 26.08 \\
Door 88 & Hybrid & 20 & 0.571 & 13.81 & 79.60 \\
\midrule
Door 150 & Raw & 4 & 0.016 & 51.28 & 109.94 \\
Door 150 & Geodesic & 44 & 0.461 & 27.11 & 41.11 \\
Door 150 & Hybrid & 20 & 0.317 & 30.95 & 73.07 \\
\midrule
Horizontal 88 & Raw & 0 & 0.052 & 47.96 & 124.53 \\
Horizontal 88 & Geodesic & 0 & 0.124 & 70.32 & 126.64 \\
Horizontal 88 & Hybrid & 0 & 0.108 & 60.90 & 123.04 \\
\bottomrule
\end{tabular}
\end{table}

\setcounter{subsection}{13}
\subsection{Temporal Horizon and Data Coverage}
\label{app:temporal_ablation_details}

This ablation varies observation coverage, pair count, and temporal separation while holding checkpoint seed 3072, the 40 distance-controlled cases, the controller settings, and the 50-action budget fixed. Table~\ref{tab:appendix_temporal_ablation} reports the full sweep underlying the main-text comparison.

Full-episode sampling draws from the available within-episode spans; in the short-horizon setting, every sampled pair has $\Delta\leq 50$. At 100,000 balanced pairs, full-episode sampling gives 97.5\% success versus 35.0\% under the shorter horizon.

\setcounter{figure}{23}
\begin{figure}[H]
\centering
\includegraphics[width=0.75\linewidth]{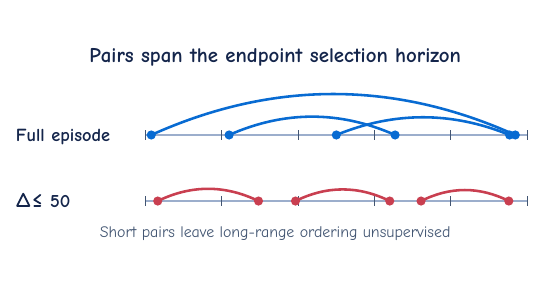}
\caption{Full-episode and short-horizon pair sampling.}
\label{fig:appendix_temporal_pair_sampling}
\end{figure}

\setcounter{table}{27}
\begin{table}[H]
\centering
\caption{TRM coverage and horizon ablations.}
\label{tab:appendix_temporal_ablation}
\begin{tabular}{@{}lrrrr@{}}
\toprule
Sampling & Rows & Pairs & Horizon & Succ. \\
\midrule
Random & 5k & 20k & Full episode & 17.5 \\
Random & 10k & 20k & Full episode & 25.0 \\
Random & 20k & 20k & Full episode & 35.0 \\
Random & 20k & 100k & Full episode & 42.5 \\
Random & 100k & 20k & Full episode & 62.5 \\
Random & 100k & 100k & Full episode & 90.0 \\
Balanced & -- & 20k & Full episode & 80.0 \\
Balanced & -- & 100k & Full episode & 97.5 \\
Balanced & -- & 20k & Max $\Delta=50$ & 5.0 \\
Balanced & -- & 100k & Max $\Delta=50$ & 35.0 \\
\bottomrule
\end{tabular}
\end{table}

\setcounter{subsection}{14}
\subsection{\pusht{} Details}
\label{app:pusht_details}

For each \pusht{} pair head, we sample 50,000 observations without replacement from the public expert demonstrations. The frozen world model encodes the pixels and proprioception in each observation. We then draw 200,000 training pairs and 10,000 validation pairs from this fixed set with independent pair seeds. The head has two hidden layers of width 256 and trains for 20 epochs with batch size 1024. We divide each target by 100 during training and retain the checkpoint with the lowest validation loss.

Closed loop evaluation uses two action dimensions. Each candidate action sequence contains five model steps and five environment actions. The action block also contains five actions, so the environment executes all five actions from the final \cem{} mean plan before the controller replans. In each iteration, \cem{} generates 300 candidate action sequences and retains the 30 candidate action sequences with the lowest terminal costs. \cem{} runs for 30 iterations with unit variance scale. After the final iteration, the controller uses the final \cem{} mean as an action sequence, and the environment executes its five actions. Goal offsets 25 and 50 use 50 episodes with an action budget of 50. Goal offset 75 uses 50 episodes per seed with an action budget of 75. Offset 25 is evaluated at seed 3072 only and has no significance test. The offset 50 and 75 results use evaluation seeds 3072--3074, with 50 episodes per seed. The fixed-pool audits use 96 candidates per episode at both offsets; the additional \cem{} comparison uses 64 candidates, 10 iterations, and common random numbers.

Let $p_o$ be object position in pixels, $p_e$ end effector position in pixels, and $\theta$ object angle in radians. The pair label is

\begin{equation}
\begin{aligned}
d_{\mathrm{task}}^2 &= \|p_o-p_o'\|_2^2+\|p_e-p_e'\|_2^2\\
&\quad+\left(\frac{180}{\pi}\,|\operatorname{wrap}(\theta-\theta')|\right)^2.
\end{aligned}
\end{equation}

Both position terms have unit weight. We convert the angle from radians to degrees before combining the three terms. We do not apply additional state normalization. We call the head trained on this distance the task state pair head. The shuffled pair head uses the same architecture and state pairs but permutes the training labels. The replacement terminal cost uses the corresponding pair head to assign a cost to each candidate action sequence. The task state hybrid standardizes the raw latent MSE cost and task state pair cost over the same candidate pool before adding them. The shuffled hybrid applies the same formula with the shuffled pair head. Threshold success records whether the environment reaches its success condition at any time. Final state success is computed from the state reached after the last executed action. It requires the joint four dimensional position error to be below 20 pixels and the wrapped angle error to be below $\pi/9$ radians.

For offsets 50 and 75, the \sasc{} tables below report ordering and selected final distance on simulator-executed candidate sequences (Appendix~\ref{app:sasc_audit}).

Compared with raw latent MSE, the task state hybrid produces a larger reduction in the final task distance of the selected candidate action sequence than an increase in binary success. At offset 75, few candidates in the 96-candidate pools meet the success threshold.

\setcounter{table}{28}
\begin{table}[H]
\centering
\caption{\pusht{} closed-loop success at goal offset 25.}
\label{tab:appendix_pusht_go25}
\begin{tabular}{lr}
\toprule
Goal offset 25 terminal cost & Success \\
\midrule
Raw latent MSE & 88.0 \\
Task state pair head & 78.0 \\
Task state hybrid & 86.0 \\
Shuffled pair head & 64.0 \\
Shuffled hybrid & 84.0 \\
\bottomrule
\end{tabular}
\end{table}

\setcounter{table}{29}
\begin{table}[H]
\centering
\caption{\pusht{} closed-loop success by seed at goal offset 50.}
\label{tab:appendix_pusht_go50_per_seed}
\begin{tabular}{@{}rrrrrr@{}}
\toprule
Seed & Raw & Pair & Hybrid & Shuff. pair & Shuff. hybrid \\
\midrule
3072 & 40.0 & 44.0 & 60.0 & 26.0 & 38.0 \\
3073 & 38.0 & 40.0 & 48.0 & 28.0 & 40.0 \\
3074 & 42.0 & 42.0 & 50.0 & 20.0 & 50.0 \\
\bottomrule
\end{tabular}
\end{table}

\setcounter{table}{30}
\begin{table}[H]
\centering
\caption{\pusht{} closed-loop results at goal offset 75.}
\label{tab:appendix_pusht_go75}
\begin{tabular}{@{}lrrrr@{}}
\toprule
Cost & $\lambda$ & Succ. & Mean dist. & Final succ. \\
\midrule
Raw & -- & 16.0 & 219.7 & 4.7 \\
Task hybrid & 0.50 & 20.7 & 149.6 & 9.3 \\
Task hybrid & 1.00 & 22.0 & 103.7 & 12.7 \\
Shuffled & 0.50 & 16.7 & 215.9 & 6.7 \\
Shuffled & 1.00 & 17.3 & 215.8 & 6.7 \\
\bottomrule
\end{tabular}
\end{table}

\setcounter{figure}{24}
\begin{figure}[H]
\centering
\includegraphics[width=1.00\linewidth]{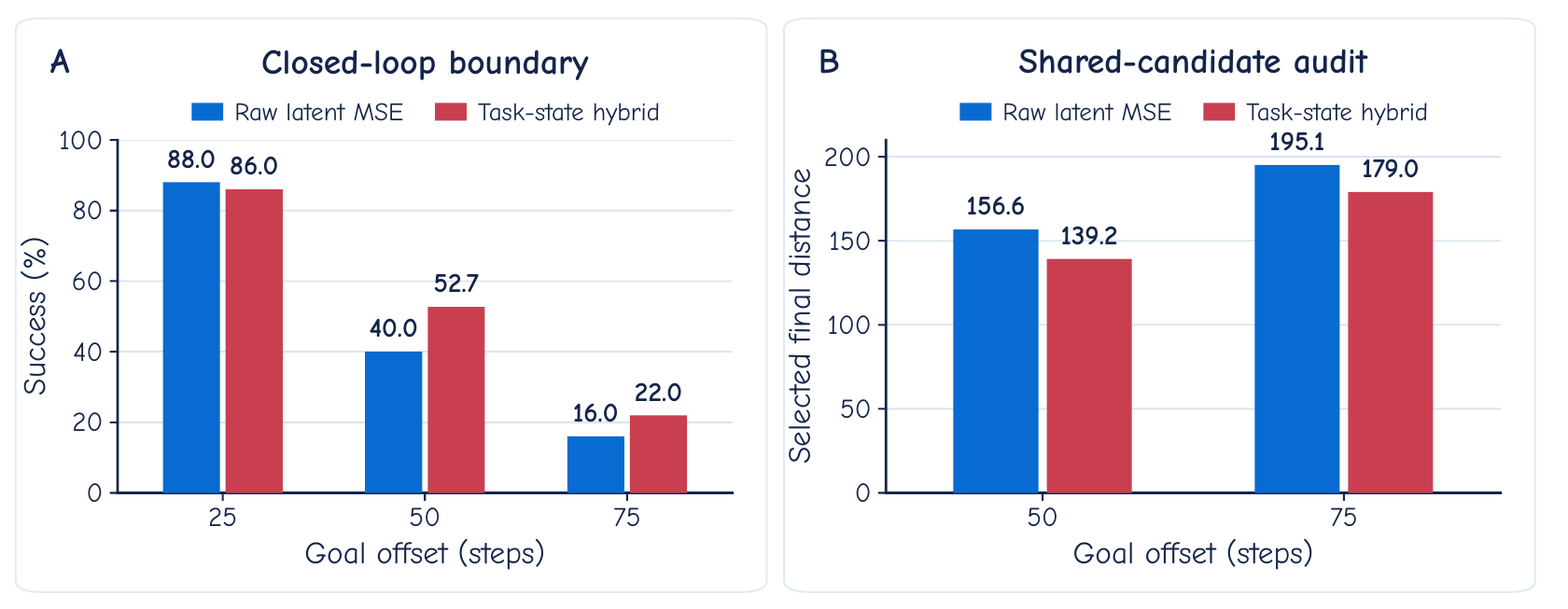}
\caption{\pusht{} closed-loop and shared-candidate diagnostics.}
\label{fig:appendix_pusht_boundary}
\end{figure}

\setcounter{table}{31}
\begin{table}[H]
\centering
\caption{\pusht{} offset 50 hybrid weight and head initialization sweep.}
\label{tab:appendix_pusht_weight_head}
\begin{tabular}{@{}llrr@{}}
\toprule
Head & Cost & $\lambda$ & Succ. \\
\midrule
None & Raw & -- & 40.0 \\
Task state & Task hybrid & 0.25 & 44.7 \\
Task state & Task hybrid & 0.50 & 46.7 \\
Task state & Task hybrid & 0.75 & 48.0 \\
Task state & Task hybrid & 1.00 & 52.7 \\
Shuffled & Shuffled hybrid & 0.25 & 42.7 \\
Shuffled & Shuffled hybrid & 0.50 & 38.7 \\
Shuffled & Shuffled hybrid & 0.75 & 42.7 \\
Shuffled & Shuffled hybrid & 1.00 & 42.7 \\
Task state, 4072 & Task hybrid & 0.50 & 48.0 \\
Task state, 4072 & Task hybrid & 1.00 & 50.0 \\
Shuffled, 4072 & Shuffled hybrid & 0.50 & 42.7 \\
Shuffled, 4072 & Shuffled hybrid & 1.00 & 43.3 \\
\bottomrule
\end{tabular}
\end{table}

\setcounter{table}{32}
\begin{table}[H]
\centering
\caption{\pusht{} offset 50 shared-candidate ranking (96 candidates).}
\label{tab:appendix_pusht_candidate}
\begin{tabular}{@{}lrrr@{}}
\toprule
Cost & Spearman & Oracle rank & Sel. dist. \\
\midrule
Raw & 0.490 & 20.41 & 156.6 \\
Task pair & 0.957 & 2.08 & 127.6 \\
Task hybrid, $\lambda=.5$ & 0.703 & 12.96 & 144.8 \\
Task hybrid, $\lambda=1$ & 0.824 & 8.37 & 139.2 \\
Shuffled pair & 0.071 & 44.86 & 175.8 \\
Shuffled hybrid, $\lambda=.5$ & 0.438 & 22.67 & 156.7 \\
Shuffled hybrid, $\lambda=1$ & 0.343 & 27.88 & 159.3 \\
\bottomrule
\end{tabular}
\end{table}

\setcounter{table}{33}
\begin{table}[H]
\centering
\caption{\pusht{} offset 75 shared-candidate ranking (96 candidates).}
\label{tab:appendix_pusht_go75_candidate}
\begin{tabular}{@{}lrrr@{}}
\toprule
Cost & Spearman & Oracle rank & Sel. dist. \\
\midrule
Raw & 0.490 & 25.73 & 195.1 \\
Task pair & 0.941 & 3.29 & 165.7 \\
Task hybrid, $\lambda=.5$ & 0.696 & 16.64 & 184.0 \\
Task hybrid, $\lambda=1$ & 0.811 & 11.07 & 179.0 \\
Shuffled pair & 0.118 & 46.58 & 216.7 \\
Shuffled hybrid, $\lambda=.5$ & 0.444 & 26.50 & 198.3 \\
Shuffled hybrid, $\lambda=1$ & 0.350 & 31.25 & 198.8 \\
\bottomrule
\end{tabular}
\end{table}

In the 64-candidate comparisons, Refined is the executed final \cem{} mean, Pool best uses the best simulator outcome as a reference, and Selected is the candidate with the lowest planner cost. Mean and median final distances refer to the state reached after the last action, separately from whether the success threshold was crossed at any earlier step.

\setcounter{table}{34}
\begin{table}[H]
\centering
\caption{\pusht{} offset 50 refinement and candidate selection (64 candidates).}
\label{tab:appendix_pusht_cem_selected}
\begin{tabular}{@{}lrrrr@{}}
\toprule
Cost & Refined & Pool best & Sel. dist. & Sel. succ. \\
\midrule
Raw & 127.4 & 103.7 & 127.3 & 16.7 \\
Task hybrid, $\lambda=1$ & 75.3 & 59.1 & 75.3 & 16.0 \\
Shuffled hybrid, $\lambda=1$ & 137.8 & 113.2 & 137.8 & 14.0 \\
\bottomrule
\end{tabular}
\end{table}

\setcounter{table}{35}
\begin{table}[H]
\centering
\caption{\pusht{} offset 75 candidate selection (64 candidates).}
\label{tab:appendix_pusht_go75_selected}
\begin{tabular}{@{}lrrrrr@{}}
\toprule
Cost & Sel. dist. & Final & Thresh. & Rank & Pool best \\
\midrule
Raw & 191.6 & 4.0 & 5.3 & 45.24 & 163.0 \\
Task hybrid, $\lambda=1$ & 114.5 & 4.0 & 4.0 & 34.63 & 96.7 \\
Shuffled hybrid, $\lambda=1$ & 183.4 & 4.0 & 4.0 & 43.04 & 158.3 \\
\bottomrule
\end{tabular}
\end{table}

\setcounter{table}{36}
\begin{table}[H]
\centering
\caption{\pusht{} offset 50 final-distance diagnostic for seed 3072.}
\label{tab:appendix_pusht_final_distance}
\begin{tabular}{@{}lrrr@{}}
\toprule
Cost & Mean & Median & Final \\
\midrule
Raw & 131.5 & 40.8 & 28.0 \\
Task hybrid, $\lambda=1$ & 48.0 & 26.2 & 42.0 \\
Shuffled hybrid, $\lambda=1$ & 154.9 & 50.9 & 30.0 \\
\bottomrule
\end{tabular}
\end{table}

\section{Execution Diagnostics and Transfer Boundaries}
\label{app:execution_diagnostics}

In the representative high-distance \tworoom{} trace, raw latent MSE favors predictions near the wall over a route through the doorway. The native \cem{} controller executes the configured 25-action prefix before replanning. Across the frozen LeWM high-distance evaluations, raw latent MSE ends in the wrong room in 45.3\% of episodes and at the wall in 10.3\% (per-checkpoint breakdown in Table~\ref{tab:appendix_hardn100_per_seed}). Shifted-door and rotated-wall results are reported in Appendix~\ref{app:coverage_layout_checks}; the contact-execution comparison is in Appendix~\ref{app:pusht_details}.

\section{Checks on the Terminal-Cost Interpretation}
\label{app:terminal_cost_checks}

The interpretation makes separate, falsifiable predictions: raw latent MSE should rank low-oracle-cost candidates poorly in an identical pool (Appendix~\ref{app:sasc_audit}); position decoded by the probe should support control when used in the terminal cost, even with episode-disjoint probe fitting (Tables~\ref{tab:appendix_subspace_per_seed} and~\ref{tab:episode_disjoint_xy}); and shuffling temporal labels should not preserve the repair (Appendix~\ref{app:tworoom_nooverlap}). Increasing the raw planning budget tests whether search volume closes the gap (Table~\ref{tab:appendix_solver_budget}). In the geodesic-head coverage intervention, lowering the validation error alone would not establish repair: oracle rank and closed-loop success must also improve (Appendix~\ref{app:coverage_layout_checks}).

\section{Additional Checkpoint Comparison}
\label{app:additional_checkpoints}

We compare seven archived \lewm{} checkpoints on a separate public TwoRooms dataset with 9,000 training and 1,000 evaluation episodes. Each frozen checkpoint has its own freshly trained temporal head. The comparison uses a common 50-episode screening set from the evaluation split, containing 34 same-room and 16 cross-wall cases, with goal offset 25 and an action budget of 50. The checkpoint identifiers in Table~\ref{tab:additional_checkpoints} retain their recorded run and epoch indices. Gain is \trm{} minus raw success in percentage points, with a 95\% paired episode-bootstrap interval. Final Euclidean distance is in pixels.

\cem{} uses 300 candidates, 30 optimization iterations, top-k 30, and solver seed 42. Planning horizon, receding horizon, and action block are each five. Each head is trained for 30 epochs with hidden dimension 256, 100,000 training pairs, and 10,000 validation pairs from the training split. Action and proprioceptive normalization are fitted on the training split.

Four checkpoints improve and two tie. All six paired intervals include zero. The epoch-3 checkpoint instead loses 30 percentage points, with interval $[-46,-14]$. These checkpoint-by-checkpoint results bound the scope of the screening comparison.

\setcounter{table}{37}
\begin{table}[H]
\caption{Seven-checkpoint screening comparison.}
\label{tab:additional_checkpoints}
\centering
\begin{tabular}{ccrrcrr}
\toprule
Run & Epoch & \shortstack{Raw\\success} & \shortstack{\trm{}\\success} & \shortstack{Gain\\{[95\% interval]}} & \shortstack{Raw\\distance} & \shortstack{\trm{}\\distance} \\
\midrule
0 & 99 & 43/50 & 48/50 & $+10\;[-2,22]$ & 16.68 & 7.91 \\
1 & 55 & 48/50 & 49/50 & $+2\;[-4,8]$ & 7.09 & 7.81 \\
2 & 3 & 45/50 & 30/50 & $-30\;[-46,-14]$ & 10.50 & 29.13 \\
3 & 19 & 41/50 & 45/50 & $+8\;[-6,22]$ & 17.40 & 14.21 \\
4 & 67 & 46/50 & 46/50 & $0\;[-10,10]$ & 11.26 & 11.04 \\
4 & 99 & 44/50 & 47/50 & $+6\;[-6,18]$ & 11.95 & 8.80 \\
5 & 99 & 46/50 & 46/50 & $0\;[-10,10]$ & 9.06 & 9.60 \\
\bottomrule
\end{tabular}
\end{table}

\section{Candidate Availability and Ranking}
\label{app:candidate_decomposition}

\textbf{Protocol.} The replay study uses the three main \lewm{} policy seeds, 3072, 3073, and 3074, with the matched high distance 100-episode evaluation set and evaluation/solver seed 44. For each policy seed, we record both the raw-cost and \trm{} controllers. \cem{} uses 300 candidates, 10 optimization iterations, top-k 30, planning and receding horizons of five, and action block five. Each episode has two recorded replanning queries. Every final-iteration candidate is executed for its complete 25-frame sequence in the \tworoom{} simulator, from the corresponding query state. In total, the comparison contains 1,200 query states and 360,000 candidate outcomes.

Both scores are evaluated on the same candidates and simulator outcomes within each pool. Let $d_0$ be the diagnostic geodesic distance at the query state and $d_i$ the distance after executing candidate $i$. Available progress is how much closer the best candidate gets; ranking regret is how much less progress the selected candidate makes. If the starting distance is 100 pixels, the pool best ends 40 pixels away, and the chosen candidate ends 65 pixels away, the available progress is 60 pixels and the regret is 25 pixels. For a selected candidate $i_s$, these quantities are

\begin{equation}
P=d_0-\min_i d_i,\qquad R_s=d_{i_s}-\min_i d_i.
\end{equation}

The diagnostic categories use $\delta=16$ pixels: generation limited means $P\leq\delta$; ranking limited means $P>\delta$ and $R_s>\delta$; near-best selection means $P>\delta$ and $R_s\leq\delta$. These are query-level progress and regret categories. Closed loop success uses the simulator's goal-hit criterion. The ranking comparison uses top-one selectors; the native controller executes its final \cem{} mean. Each pool source contributes 600 replanning states. The three category percentages use all 600 states; mean regret uses 599 finite outcomes for raw scoring on raw pools and 600 for each other row.

\setcounter{table}{38}
\begin{table}[H]
\caption{Candidate availability and ranking under shared simulator outcomes.}
\label{tab:candidate_decomposition}
\centering
\begin{tabular}{llrrrr}
\toprule
Pool source & Selector & \shortstack{Generation\\limited} & \shortstack{Ranking\\limited} & \shortstack{Near-best\\selection} & \shortstack{Mean\\regret} \\
\midrule
Raw & Raw & 26.67 & 61.33 & 12.00 & 22.67 \\
Raw & \trm{} & 26.67 & 28.00 & 45.33 & 13.02 \\
\trm{} & Raw & 0.33 & 26.83 & 72.83 & 13.32 \\
\trm{} & \trm{} & 0.33 & 31.83 & 67.83 & 8.69 \\
\bottomrule
\end{tabular}
\end{table}

Across the paired replay states, mean finite regret falls with \trm{} scoring in both pool sources (Table~\ref{tab:candidate_decomposition}). This does not imply improvement at every query: on \trm{}-generated pools the ranking-limited share rises from 26.83\% to 31.83\%. The two generators also visit different query states.

\textbf{Native execution and remaining failures.} Native \trm{} succeeds on 291/300 episodes versus 21/300 for raw cost. Its nine failures contain 18 replanning states: zero generation-limited, five ranking-limited, and thirteen with near-best top-one selection.

\textbf{Oracle selection within the pool.} A separate closed loop run keeps each native candidate generator and selects the candidate with the smallest simulator-evaluated terminal geodesic distance. It executes that candidate's complete 25-frame sequence and replans. Table~\ref{tab:within_pool_oracle} reports the results. Native success records reaching the goal at any time during execution; the oracle's selection objective uses the endpoint after the full sequence.

\setcounter{table}{39}
\begin{table}[H]
\caption{Closed-loop outcomes with simulator oracle selection.}
\label{tab:within_pool_oracle}
\centering
\begin{tabular}{lrrrrr}
\toprule
Pool generator & Seed 3072 & Seed 3073 & Seed 3074 & Total & Success (\%) \\
\midrule
Raw & 28 & 24 & 10 & 62/300 & 20.7 \\
\trm{} & 99 & 100 & 99 & 298/300 & 99.3 \\
\bottomrule
\end{tabular}
\end{table}

\textbf{Candidate-budget comparison.} We compare 50, 100, 300, and 1,000 candidates at the same 100 initial states for each policy seed and generator, totaling 2,400 state--arm observations and 870,000 candidate outcomes. The 300-candidate outcomes reuse the first-query replay above. Table~\ref{tab:candidate_budget_additional} reports available progress, regret over finite results, and the percentage of candidates with unreachable endpoints. More candidates increase available progress for both generators. On these initial states, lowering regret is a separate outcome: raw scoring has slightly lower regret on \trm{}-generated pools at 100, 300, and 1,000 candidates.

\setcounter{table}{40}
\begin{table}[H]
\caption{Initial-state candidate-budget comparison across three policy seeds.}
\label{tab:candidate_budget_additional}
\centering
\begin{tabular}{rlrrrr}
\toprule
Candidates & Generator & \shortstack{Available\\progress} & \shortstack{Raw\\regret} & \shortstack{\trm{}\\regret} & \shortstack{Unreachable\\(\%)} \\
\midrule
50 & Raw & 16.80 & 7.88 & 5.83 & 0.240 \\
50 & \trm{} & 92.09 & 10.08 & 9.78 & 0.640 \\
100 & Raw & 26.82 & 15.36 & 10.23 & 0.137 \\
100 & \trm{} & 111.60 & 15.50 & 16.00 & 0.987 \\
300 & Raw & 38.17 & 24.01 & 14.25 & 0.221 \\
300 & \trm{} & 127.83 & 20.30 & 21.88 & 0.400 \\
1000 & Raw & 45.63 & 30.19 & 16.67 & 0.235 \\
1000 & \trm{} & 132.25 & 23.00 & 23.82 & 0.289 \\
\bottomrule
\end{tabular}
\end{table}

\section{Separate Temporal-Only Diagnostic}
\label{app:new_continuous}

This section records a separate temporal-only diagnostic, not the per-seed source of the main continuous-control comparison in Table~\ref{tab:new_continuous}.

\textbf{Protocol and statistical unit.} Reacher and Cube each use a frozen LeWorldModel epoch-10 checkpoint trained with seed 6072. Seeds 3072, 3073, and 3074 vary the temporal head and/or solver, not the world model. Each evaluation uses 100 episodes, an initial state at frame 0, and a goal at frame 25. Control uses a 50-step budget, planning horizon 5, receding horizon 5, action block 5, 300 samples, 30 \cem{} iterations, and 30 elites. The head uses an 800/100/100 episode split and the world model uses a 0.9 training split. The records establish fixed-model scoring comparisons but do not verify that head-test trajectories were unseen in world-model training.

\setcounter{table}{41}
\begin{table}[H]
\caption{Per-seed success counts for the separate temporal-only diagnostic.}
\label{tab:new_continuous_seeds}
\centering
\begin{tabular}{llrrrr}
\toprule
Task & Terminal cost & Seed 3072 & Seed 3073 & Seed 3074 & Mean (\%) \\
\midrule
Reacher & Raw latent & 89 & 88 & 89 & 88.67 \\
Reacher & Temporal TRM, width 256 & 79 & 75 & 73 & 75.67 \\
Reacher & Temporal TRM, width 1024 & 78 & 83 & 76 & 79.00 \\
\midrule
Cube & Raw latent & 6 & 6 & 8 & 6.67 \\
Cube & Temporal TRM, width 256 & 6 & 4 & 6 & 5.33 \\
Cube & Temporal TRM, width 1024 & 7 & 7 & 8 & 7.33 \\
\bottomrule
\end{tabular}
\end{table}

The width-1024 head outperforms width 256 on both tasks in mean success. Both heads still underperform raw latent MSE on Reacher, while all Cube means remain below 8\% (Table~\ref{tab:new_continuous_seeds}). Thus width changes the temporal-only result without producing consistent gains over raw cost.

\section{Direct Goal-Conditioned Policy References}
\label{app:direct_rl}

\textbf{Offline goal-conditioned IQL.} The TwoRoom comparison uses three corresponding frozen world models with seeds 3072, 3073, and 3074 and 192-dimensional latent observations. The actor receives three latent frames and the goal representation and executes a single action without planning. Training uses the 9,807 eligible logged trajectories after excluding evaluation-source trajectories from head training. This exclusion does not establish that the world models were trained without those trajectories. The critic is trained for 100,000 updates and then the actor for 100,000 updates, each with batch size 32 (3.2 million sampled windows per phase). No online interaction is used for training. Unlike \trm{} pair regression, IQL also consumes logged actions; neither the number of training samples nor the training compute is matched to \trm{}.

Evaluation uses the strict high-distance 100-case protocol and a 50-action budget. Success is the logical OR of reaching within a strictly less than 16-pixel distance threshold during execution. The three IQL runs share the same evaluation manifest, and none of their cases is initially successful. For each seed, all 100 cases also match the raw, episode-excluded temporal, and episode-excluded shuffled \trm{} records on start and goal rows, state identifiers, and XY coordinates. The corresponding temporal \trm{} counts are 99/96/95, shuffled counts are 0/0/0, and raw counts are 16/5/0. Recomputing success as minimum executed distance strictly below 16 pixels agrees with every recorded success entry in all nine original comparison arms (900 cases). This establishes matched tasks and the same observed success criterion, not identical evaluator source code or matched training compute. The successes are 78, 98, and 78 out of 100, giving 84.67\%. This reference compares direct policy learning with model-based planning; it does not isolate a Q-based terminal score inside the same \cem{} controller.

\textbf{Online PPO.} The online reference uses Stable-Baselines3 PPO 2.7.1. Each completed run trains for 5,242,880 environment steps with 32 environments, 64 rollout steps per environment, batch size 256, 10 optimization epochs per rollout, and a two-layer width-256 policy network. The records contain 204,800 optimizer steps. We evaluate the final checkpoint rather than selecting a model on development performance; development evaluation is diagnostic only.

The observations contain two frames of frozen LeWorldModel latents, the goal latent, and their current--goal difference, using world-model seed 6072. Physical state is not an actor input, but it supplies the dense training reward. This additional reward supervision and online experience distinguish PPO from the offline temporal-head experiment.

Each test contains 100 cases, the goal at an offset of 25 frames, and a 50-step control budget. PPO and all three \cem{} score variants use identical initial data-row indices, with the same goal-row rule. The legacy \cem{} records do not include initial physical-state hashes, so matching these data rows is not a verification of bitwise-identical simulator resets.

\setcounter{table}{42}
\begin{table}[H]
\caption{Direct-policy results and initial-success accounting.}
\label{tab:direct_rl_seeds}
\centering
\begin{tabular}{llrrrr}
\toprule
Task & Policy & Seed 3072 & Seed 3073 & Seed 3074 & Mean (\%) \\
\midrule
TwoRoom & Offline GC-IQL & 78 & 98 & 78 & 84.67 \\
Reacher & Online PPO & 3 & 6 & 3 & 4.00 \\
Cube & Online PPO & -- & -- & 4 & 4.00 (one seed) \\
\bottomrule
\end{tabular}
\end{table}

Every Reacher PPO case starts outside success. In the completed Cube evaluation (seed 3074), all four successes are already successful at initialization; none of the other 96 cases newly succeeds. The remaining two Cube evaluations were unavailable due to framebuffer failures. Table~\ref{tab:direct_rl_seeds} therefore reports one valid Cube seed. Dashes mark unavailable evaluations, not zero success.

\section{Pairwise Head Settings}
\label{app:head_settings}

Section~\ref{app:method_training} specifies the head and training procedure. Shuffled controls keep the architecture and state pairs fixed and permute the training labels.

\setcounter{subsection}{1}
\subsection{Implementation Details for the Experimental Comparisons}
\label{app:experimental_implementation}

Table~\ref{tab:appendix_controller_config} specifies the default \tworoom{} planner. The single-seed ranking diagnostic uses checkpoint 3072 (Table~\ref{tab:appendix_hard_sasc}). Section~\ref{app:candidate_decomposition} describes multi-seed simulator replay, and Section I gives the direct-policy training budgets.

\setcounter{subsection}{2}
\subsection{Additional Reporting Details}
\label{app:reporting_v008}

Navigation comparisons use three LeWM and three \pldm{} seeds. Coverage and pair-count sweeps use one checkpoint. Section F reports the separate checkpoint screening.

\clearpage
\section{Detailed Tables Accompanying Main-Text Figures}
\label{app:retained_tables}

These tables collect the data behind the main-text figures. The long-distance \tworoom{} success rates use the mean and sample standard deviation across three seeds; the other percentage columns show seed means. The direct-policy comparison uses the same 100 start--goal cases per seed and a 50-action budget, but supervision and optimization budgets differ. The matched metric-baseline table averages three seeds. Learned geodesic baselines use oracle supervision (Section~\ref{app:coverage_layout_checks}).

\setcounter{table}{43}
\begin{table}[H]
\caption{Core diagnostic tasks.}
\label{tab:evaluation_suite}
\centering
\begin{tabular}{L{0.12\textwidth}L{0.16\textwidth}L{0.20\textwidth}L{0.15\textwidth}L{0.21\textwidth}}
\toprule
Task & Role & Data source & Seeds & Primary evidence \\
\midrule
\tworoom{} & Main navigation task & Public \leworldmodel{} cache & 3072/3073/\allowbreak3074 & Success, \sasc{}, rowspace audits \\
TopoNav & Non-\tworoom{} navigation diagnostic & Generated grid trajectories & 3072/3073/\allowbreak3074 & Success and exact graph distance \sasc{} \\
\pusht{} & Contact rich boundary task & Public PushT expert data & 3072/3073/\allowbreak3074 & Success, selected distance, \sasc{} \\
\bottomrule
\end{tabular}
\end{table}

\setcounter{table}{44}
\begin{table}[H]
\caption{High distance \tworoom{} results across three seeds.}
\label{tab:hardn100}
\centering
\begin{tabular}{lrrrrr}
\toprule
Cost & Succ. & Same & Cross & Wrong & Stuck\\
\midrule
Raw latent MSE & $7.0{\pm}8.2$ & 10.0 & 4.0 & 45.3 & 10.3  \\
TRM, complete cache & $97.0{\pm}3.5$ & 95.3 & 98.7 & 0.0 & 0.3  \\
TRM, episode exclusion & $96.7{\pm}2.1$ & 98.7 & 94.7 & 2.0 & 0.3  \\
Shuffled temporal head & $0.0{\pm}0.0$ & 0.0 & 0.0 & 46.0 & 8.7  \\
\midrule
\pldm{} raw latent MSE & $32.7{\pm}3.2$ & 4.0 & 61.3 & 12.7 & 14.0 \\
\pldm{} temporal TRM & $84.0{\pm}9.5$ & 79.3 & 88.7 & 3.0 & 2.0\\
\pldm{} shuffled head & $0.0{\pm}0.0$ & 0.0 & 0.0 & 50.0 & 2.7 \\
\bottomrule
\end{tabular}
\end{table}

\par\noindent\begin{minipage}[t]{.48\linewidth}
\setcounter{table}{45}

\captionof{table}{TwoRoom control comparison with direct-policy learning.}
\label{tab:gc_iql}
\centering
\begin{tabular}{lr}
\toprule
Method & Success \\
\midrule
Raw latent planning & 7.00 \\
TRM, episode exclusion & 96.67 \\
Offline GC-IQL & 84.67 \\
\bottomrule
\end{tabular}

\end{minipage}\hfill
\begin{minipage}[t]{.48\linewidth}
\setcounter{table}{47}

\captionof{table}{Metric baselines on matched high distance candidate pools.}
\label{tab:decisive_stress}
\centering
\setlength{\tabcolsep}{4pt}
\begin{tabular}{lrrrr}
\toprule
Selector & Succ. & $\rho_{\mathrm{geo}}$ & Rank & Dist. \\
\midrule
Raw latent MSE & 7.0 & 0.007 & 38.95 & 139.72 \\
Standardized L2 & 3.3 & 0.007 & 40.34 & 147.59 \\
Linear & 15.3 & 0.354 & 17.81 & 88.75 \\
Diag. Mahalanobis & 21.3 & 0.279 & 18.93 & 99.37 \\
Decoded XY geodesic & 93.3 & 0.857 & 0.56 & 10.90 \\
Temporal \trm{} & 97.0 & 0.678 & 3.79 & 7.85 \\
Geodesic pair \mlp{} & 99.3 & 0.908 & 0.45 & 7.61 \\
Geodesic ranking & 95.7 & 0.839 & 0.70 & 9.22 \\
Shuffled temporal & 0.0 & 0.053 & 44.10 & 138.86 \\
Oracle geodesic & 100.0 & 1.000 & 0.00 & 5.74 \\
\bottomrule
\end{tabular}

\end{minipage}\par

\end{document}